\documentclass[10pt]{article} 
\usepackage[accepted]{rlc}

\usepackage{amssymb}            
\usepackage{mathtools}          
\usepackage{mathrsfs}           
\usepackage{graphicx}           
\usepackage{subcaption}         
\usepackage[space]{grffile}     
\usepackage{url}                
\usepackage{graphbox}

\usepackage{array}
\usepackage{booktabs}

\title{The Interpretability of Codebooks in Model-Based Reinforcement Learning is Limited}

\author{Kenneth Eaton\textsuperscript{1}, Jonathan Balloch\textsuperscript{2}, Julia Kim\textsuperscript{2}, Mark Riedl\textsuperscript{2}  \\
    \textsuperscript{1}Georgia Tech Research Institute,
    \textsuperscript{2}Georgia Institute of Technology}

\begin{document}

\maketitle

\begin{abstract}
Interpretability of deep reinforcement learning systems could assist operators with understanding how they interact with their environment.
Vector quantization methods---also called {\em codebook} methods---discretize a neural network's latent space that is often suggested to yield emergent interpretability.
We investigate whether vector quantization in fact provides interpretability in model-based reinforcement learning.
Our experiments, conducted in the reinforcement learning environment Crafter, show that the codes of vector quantization models are inconsistent, have no guarantee of uniqueness, and have a limited impact on concept disentanglement, all of which are necessary traits for interpretability. 
We share insights on why vector quantization may be fundamentally insufficient for model interpretability.
\end{abstract}

\section{Introduction}
\label{sec:introduction}
Although deep neural networks have advanced the performance of reinforcement learning (RL) agents, these ``black box'' models give little insight as to the agent's decision making and learned transition models.
Interpretability of RL agents is important in high-stakes domains that demand human trust, such as autonomous vehicles and infrastructure. Interpretability is also crucial to analyze behavior and develop adaptations when trained agents make mistakes or experience novel changes to their environment.

There has been significant recent interest in vector quantization (VQ) in neural networks, largely driven by the work of~\cite{van2017neural}; the authors show that VQ latent space discretization in generative models helps regularize the latent space, thereby improving performance. 
Although not claimed by~\cite{van2017neural}, several subsequent works suggest that the latent disentanglement caused by VQ enables greater semantic interpretability of neural models~\citep{NIPS2017_0a0a0c8a,tamkin2023codebook,aloufi2020privacy}.

We examine whether VQ within model-based reinforcement learning (MBRL) captures semantic information about entities in the environment. 
Specifically, we use the IRIS~\citep{micheli2022transformers} MBRL agent, which quantizes the latent vector encoding of the world state and uses a transformer to model transition dynamics.
Grad-CAM~\citep{selvaraju2017grad} is used to perform qualitative and quantitative analysis to determine whether codes are consistent and represent semantically groundable entities, such as objects, across diverse inputs.

We find VQ lacks the necessary constraints to enforce that disentangled codes correspond to semantic entities; codes are not consistent enough to be grounded to a semantic concept.
Based on our findings, we disprove the intuition that VQ alone is sufficient for interpretability and hypothesize
that latent semantic alignment is needed alongside discretization to make latent spaces 
generally interpretable.


\begin{figure}[t!]
    \centering
    \includegraphics[width=0.78\textwidth]{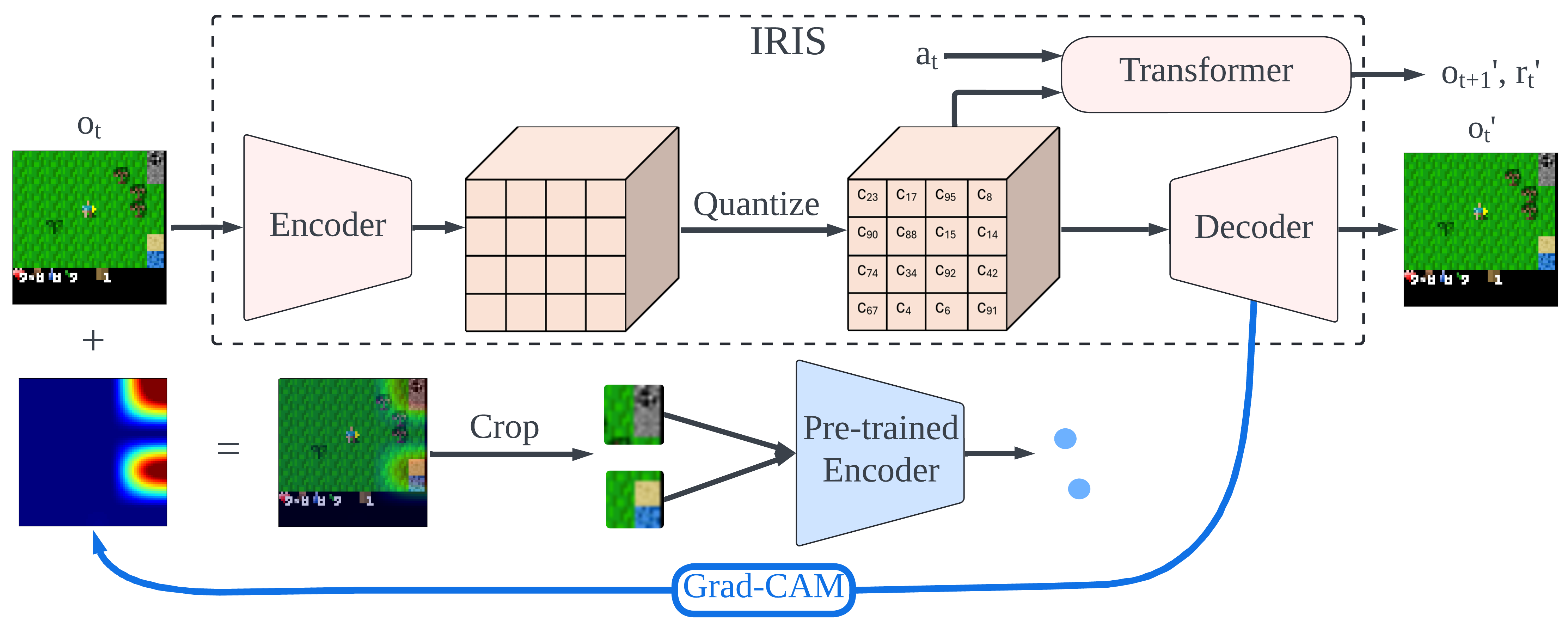}
    \caption{IRIS transition model architecture and our evaluation process. Heatmaps from Grad-CAM crop the input image to focus regions, which are then embedded by a pre-trained encoder.}
    \label{fig:arch}
\end{figure}

\section{Background and Related Work}
In MBRL, the agent learns a model to approximate the transition dynamics of the environment~\citep{sutton1998introduction}.
An agent can benefit from having a good predictive model of the future state~\citep{Werbos:87specifications}, commonly dubbed a {\em World Model}~\citep{ha2018world, hafner2020mastering}.
IRIS~\citep{micheli2022transformers} is a MBRL agent that uses VQ to replace the latent state vectors of an encoded observation with the most similar codes, which are input to a decoder and a transformer learning the World Model.
Since the codebook sits between the encoder and decoder, individual codes have the potential to represent grounded entities~\citep{khazatsky2021can}.

VQ is a data compression technique that models a latent distribution as a discrete set of vectors or ``codes.''
Formally, a VQ function $Q$ can be defined as:
   $ Q : R^d \rightarrow C $,
where the ``codebook'' $C = \{c_i|i \in I, c_i \in R^d\}$ is composed of $k$ vectors referred to as {\em codes}.
For a given input $x$, Q assigns a code $c_i$ that minimizes the MSE between the vectors.
In deep neural networks, VQ is most often used in the latent space of generative models, replacing the encoder output with learned codes~\citep{van2017neural,esser2021taming,ramesh2021zero}.
VQ is also used in the latent space of RL models to constrain and regularize them, which improves accuracy, robustness to noise, and sample efficiency~\citep{zhang2022deep, robine2020discrete}.

Several works have examined and suggested the emergent interpretability of VQ methods. 
\cite{chen2022vector,zhang2022deep} demonstrate that the entire latent embedding created from a combination of codes is capable of capturing consistent states.
There are also works that suggest the interpretability of some specific codes using qualitative assessments such as latent traversals~\citep{aloufi2020privacy,shu-etal-2020-controllable,desticourt2022topo,zou2023disentangling,tamkin2023codebook,wallingford2023neural,shao2023compositional,hsu2024disentanglement}. 
Despite evidence from works such as \cite{pmlr-v97-locatello19a}, the suggestion of VQ interpretability comes from an unsubstantiated connection between latent \textit{disentanglement} and latent semantics~\citep{guo-etal-2020-evidence,khazatsky2021can} or interpretability~\citep{NIPS2017_0a0a0c8a,LIU2022102516,zou2022joint,klein2022improving}. 
These methods rarely quantify the interpretability provided by VQ codes, and we found no prior work that 
studies VQ interpretability in MBRL. 

This work is similarly motivated to the parallel research area of \textit{mechanistic interpretability}~\cite{nanda_2022,elhage2022superposition} (MI), which studies interpreting neural network behavior based on an interpretation of the network's internal structures~\cite{olah2020zoom}. 
This stands in contrast to ``black-box''~\cite{molnar2022interpretable} interpretability approaches, such as saliency maps of inputs from outputs~\cite{adebayo2018sanity,jain2019attention,wiegreffe2019attention}, which attempt to explain relationships between the inputs and outputs without considering the network's internal mechanisms. 
Interpreting neural networks using codebooks falls in between these methods (in what is sometimes called ``white-box''~\cite{molnar2022interpretable,rauker2023toward} interpretability),  where (as in MI) the internal structures of the neural networks are constrained or investigated to establish what activations of the network are representing, while using tools like saliency maps to convey feature importance.
By taking a white-box approach to interpreting codes while using MI concepts such as superposition~\cite{elhage2022superposition}, our analysis balances the theoretical robustness of MI with the ability to interpret almost any network without the need to understand all its internal structures.

\section{Experiments in VQ Interpretability}

\paragraph{Experimental Approach.}
In our work, we use the state-of-the-art MBRL model IRIS~\citep{micheli2022transformers} featuring a codebook with 512 codes.
We make no modifications to the IRIS architecture.
To evaluate interpretability, we apply Grad-CAM to the code inputs of the decoder.
Grad-CAM scores importance of input image pixels based on the activations and gradients of a selected weight vector or layer.
We apply this process individually for each code in the model by masking out the other codes' values.
The end result is a heatmap for each code that overlays the input image with the regions that were most influential in that code being selected.

To measure the quantitative properties of the learned codes, we evaluate the code consistency over agent observations and episodes by looking at the areas of high Grad-CAM activation for the same code.
We split images into connected components of high activation and filter out those with an area smaller than a threshold.
The remaining connected components are used to crop the original images to the bounding box region they encapsulate.
We use a frozen pre-trained model to compare the cropped inputs and gauge their semantic similarity.

We conducted our experiments in  Crafter~\citep{hafner2021crafter}, an open-world survival game designed for RL with similar dynamics to Minecraft.
The model and corresponding policy was first trained to convergence.
Then we collected a dataset of 127,434 $(o_t,s_t,a_t,s_{t+1},o_{t+1})$ transitions of the agent acting in the environment over 714 episodes, where $o$ is the input observation, $a$ is the selected action, and $s$ are the latent states. 
We applied our interpretation approach to investigate different codes correspondance to the inputs on newly collected runs, and saved the Grad-CAM activation heatmaps for every code the model selected in each observation.

We found that 90\% of the heatmaps contained all zero values.
Since all the values being zero is not useful for our interpretability evaluation, those samples were dropped.
477 of the 512 codes had non-zero heatmaps, equating to 205,231 heatmaps for the final dataset.

\paragraph{Results.}
Table \ref{Tab:heatmaps} displays eight heatmaps overlaid on their observation image, randomly sampled for each code, from the most frequently occurring code in our dataset, the tenth most frequently occurring, and the median occurring.
We chose these codes based on their frequency to avoid biasing the results and capture diverse regions of the frequency distribution, as discussed further in Appendix \ref{sec:appendix1}.
In general, the heatmaps from a code focus on the same regions of the image, especially for the more frequently occurring codes.
To examine the content the codes focus on, in Table \ref{Tab:crops} we show the crops resulting from the heatmaps. To display results for more codes while still focusing on diverse frequencies, we show crops for the second most frequent code, eleventh most frequent, and the next most frequent code after the median code used earlier.
In the crops, code 46 is consistently focused on numbers indicating agent inventory, while the other two have several similar crops but no clear, singular focus.
Therefore, we conclude codes did not consistently associate with objects or concepts.
Additional examples of crops are presented in Appendix \ref{sec:appendix2}.

\begin{table}\sffamily
\centering
\caption{Example heatmaps for codes.}
\begin{tabular}{l*8{c@{\hspace{4mm}}}}
\toprule
Code 304 & 
\includegraphics[align=c, width=0.08\textwidth]{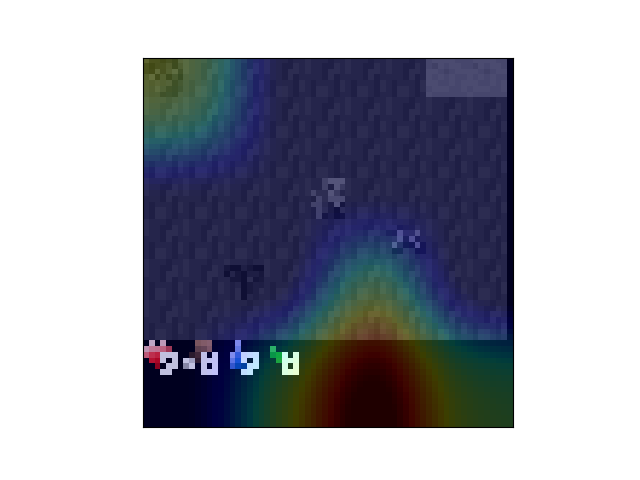} & \includegraphics[align=c, width=0.08\textwidth]{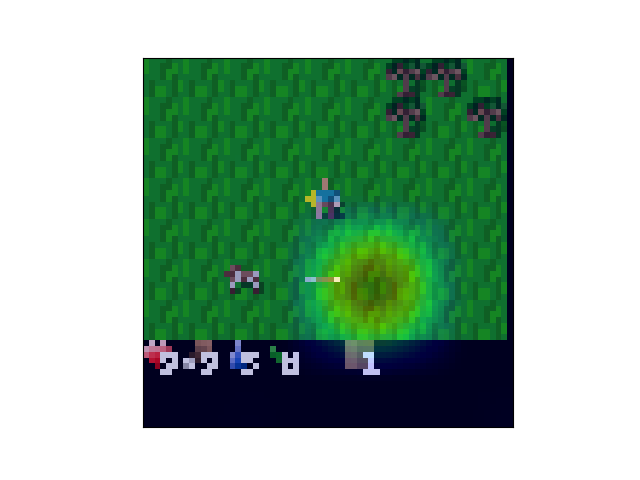} & \includegraphics[align=c, width=0.08\textwidth]{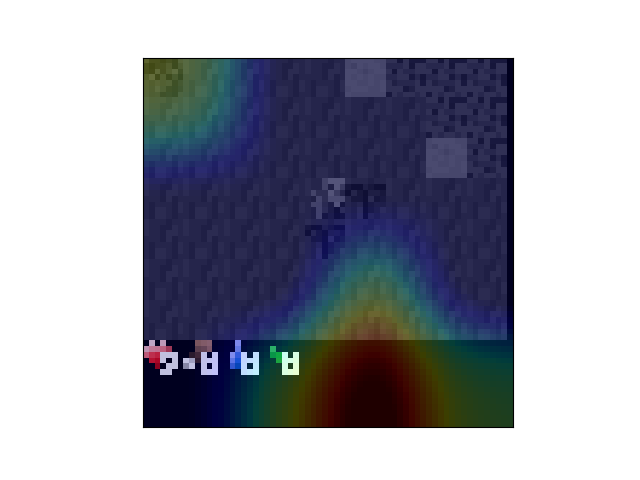} & \includegraphics[align=c, width=0.08\textwidth]{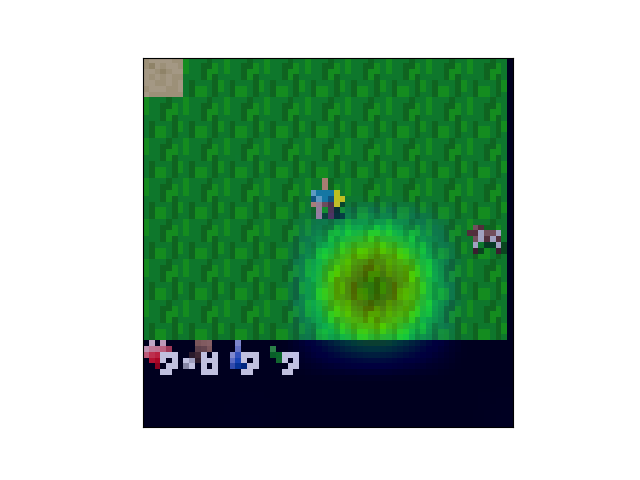} & \includegraphics[align=c, width=0.08\textwidth]{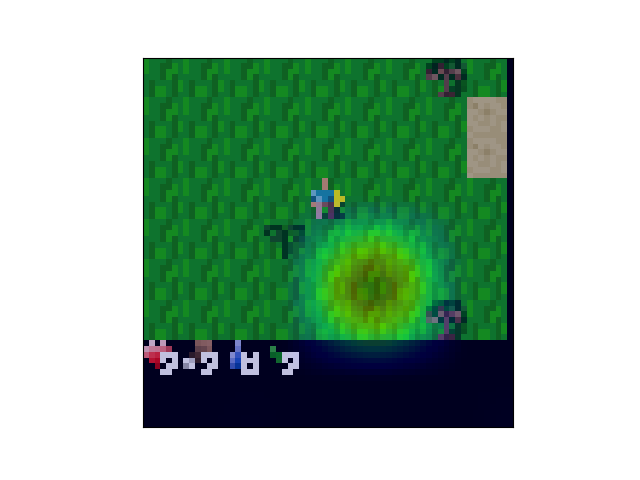} & \includegraphics[align=c, width=0.08\textwidth]{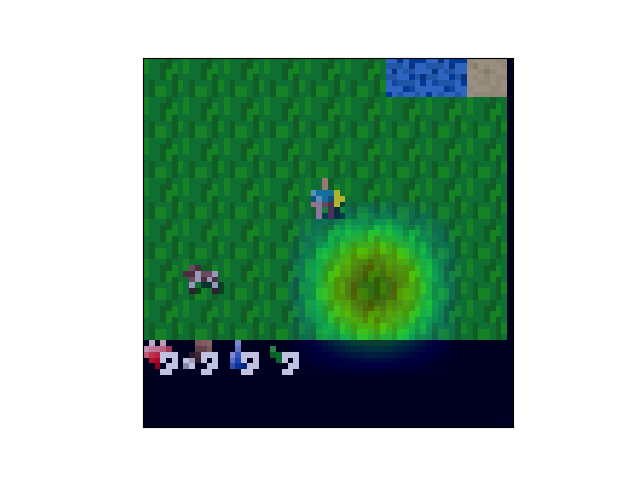} & \includegraphics[align=c, width=0.08\textwidth]{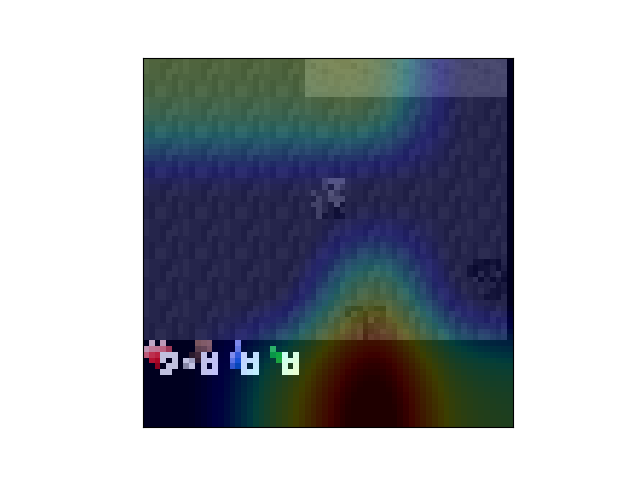} & \includegraphics[align=c, width=0.08\textwidth]{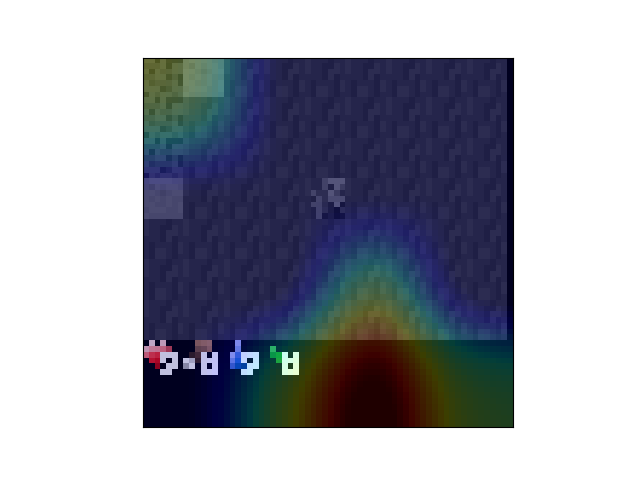} \\ 
Code 434 & 
\includegraphics[align=c, width=0.08\textwidth]{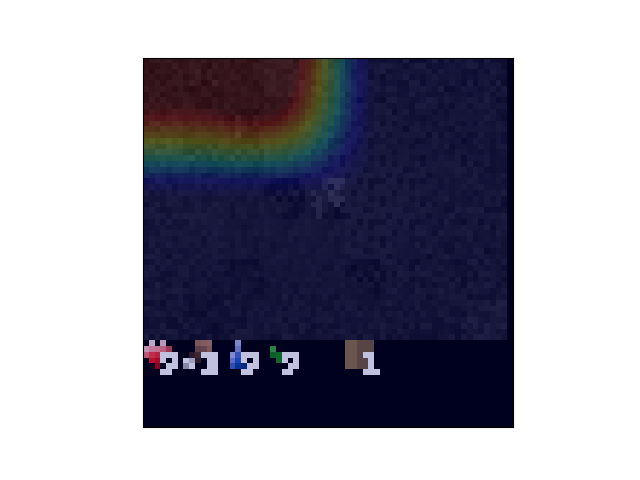} & \includegraphics[align=c, width=0.08\textwidth]{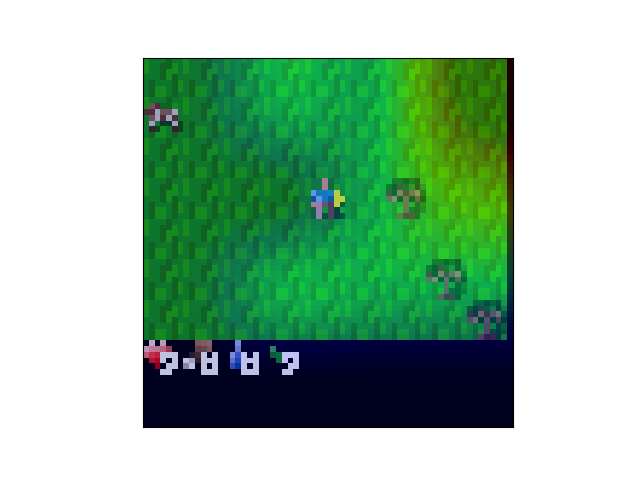} & \includegraphics[align=c, width=0.08\textwidth]{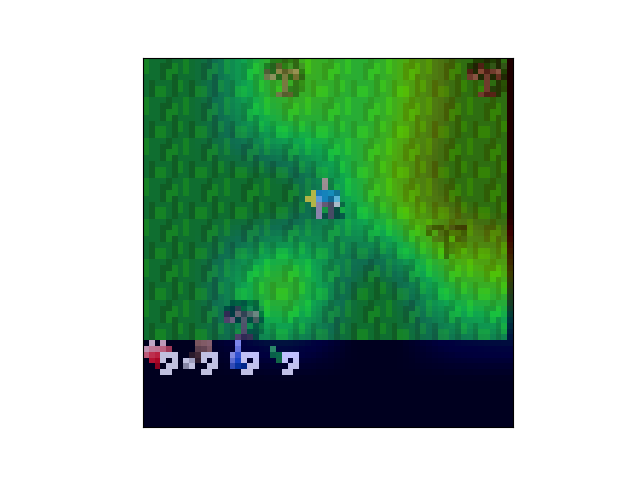} & \includegraphics[align=c, width=0.08\textwidth]{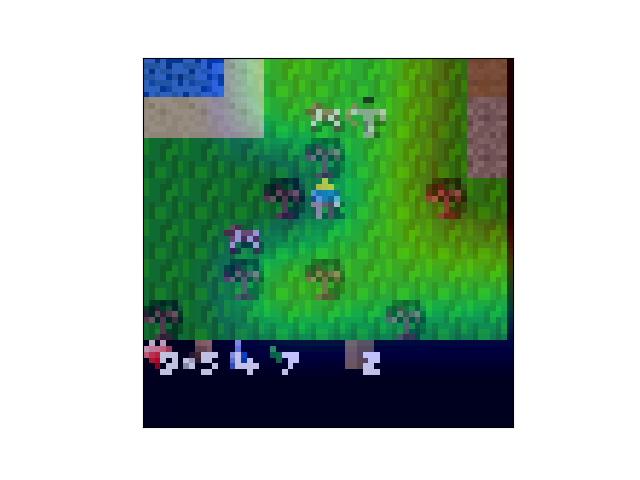} & \includegraphics[align=c, width=0.08\textwidth]{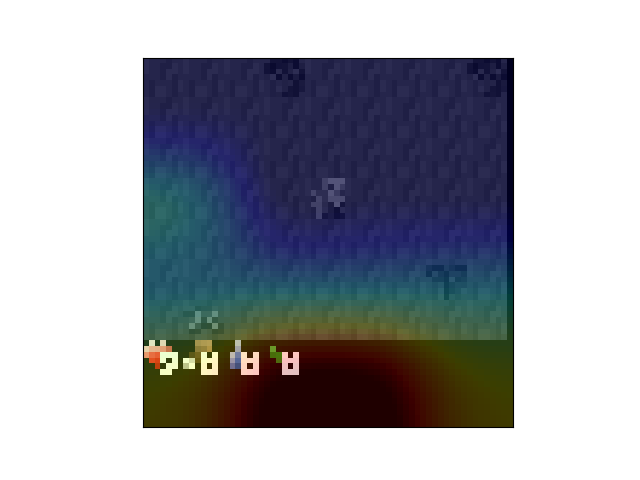} & \includegraphics[align=c, width=0.08\textwidth]{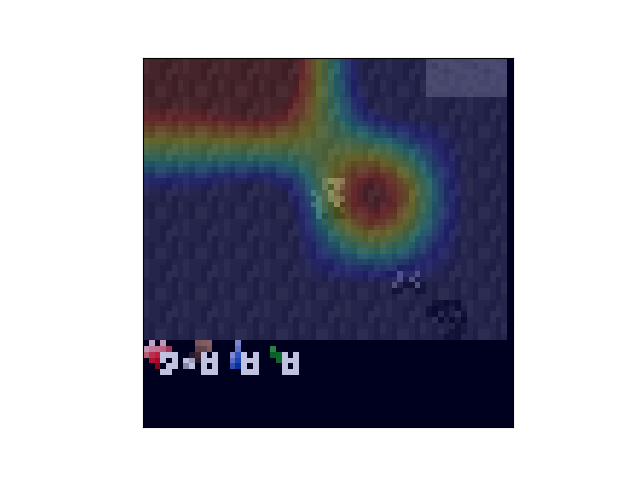} & \includegraphics[align=c, width=0.08\textwidth]{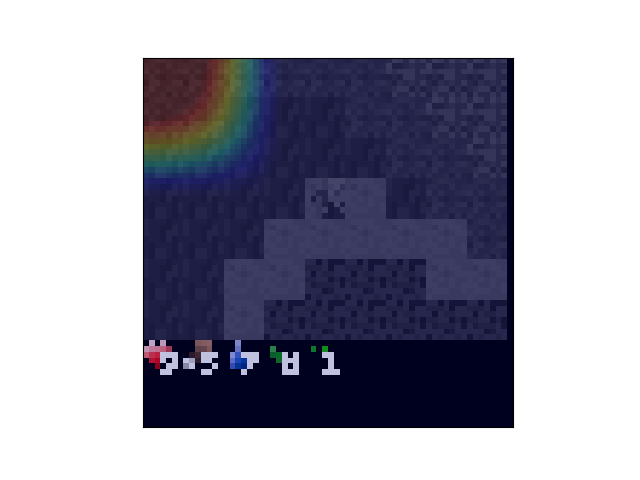} &  \includegraphics[align=c, width=0.08\textwidth]{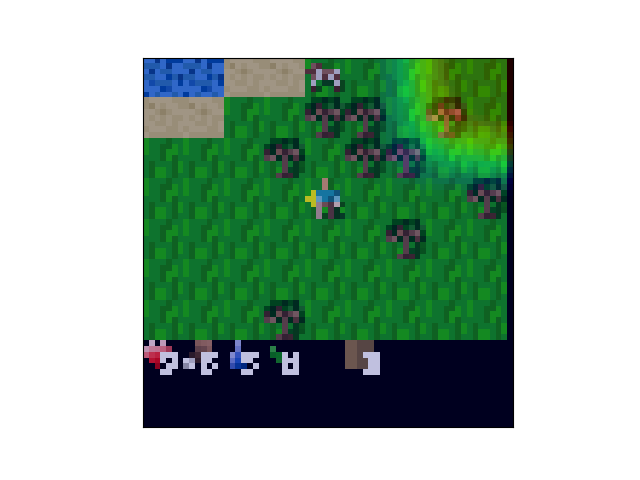} \\ 
Code 381 & 
\includegraphics[align=c, width=0.08\textwidth]{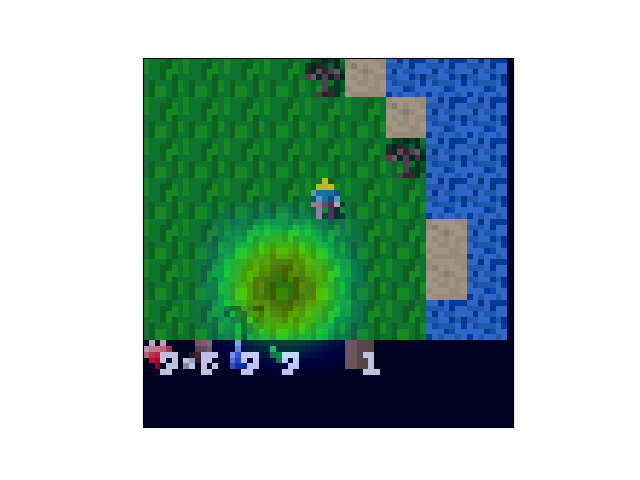} & \includegraphics[align=c, width=0.08\textwidth]{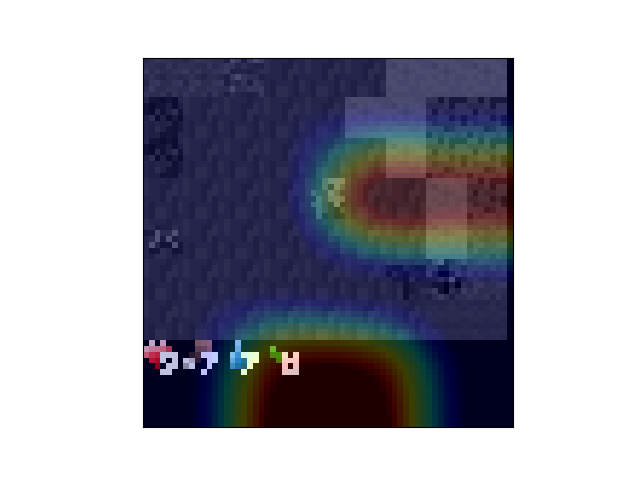} & \includegraphics[align=c, width=0.08\textwidth]{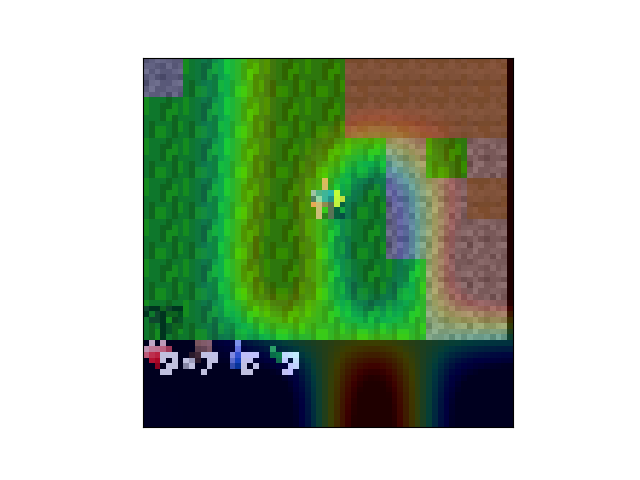} & \includegraphics[align=c, width=0.08\textwidth]{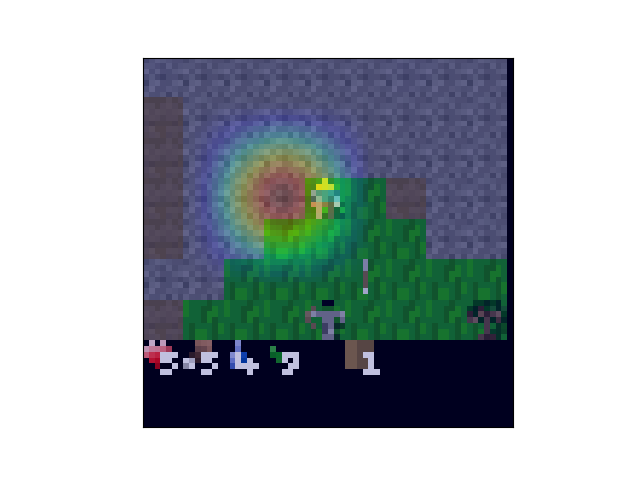} & \includegraphics[align=c, width=0.08\textwidth]{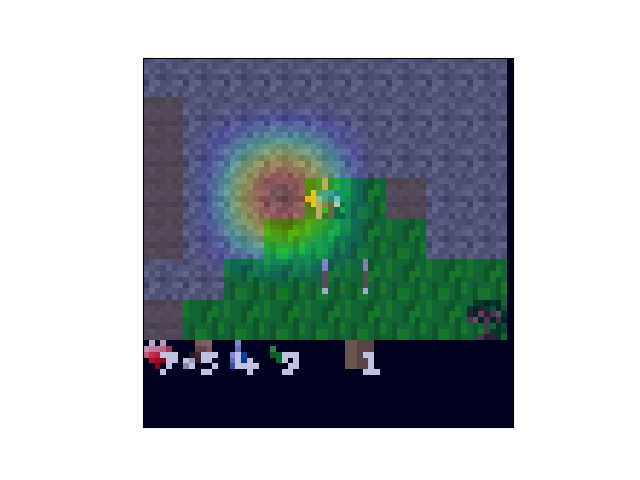} & \includegraphics[align=c, width=0.08\textwidth]{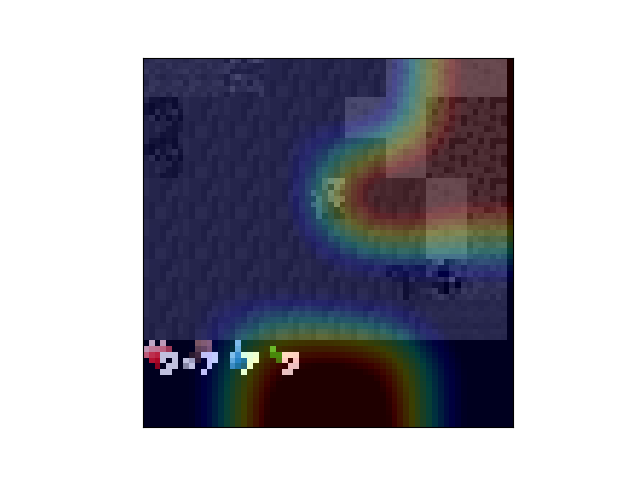} & \includegraphics[align=c, width=0.08\textwidth]{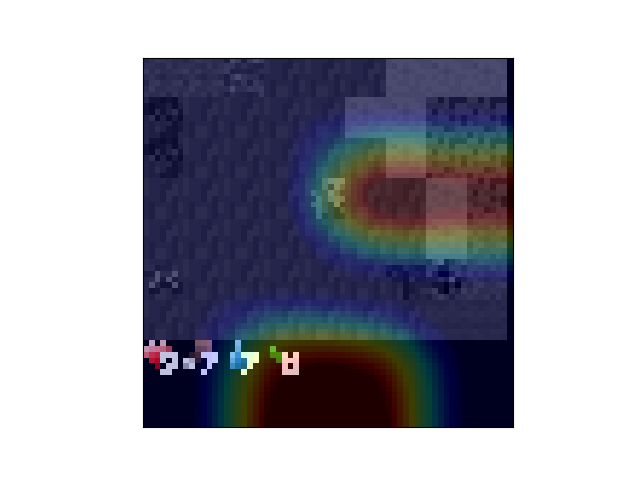} & \includegraphics[align=c, width=0.08\textwidth]{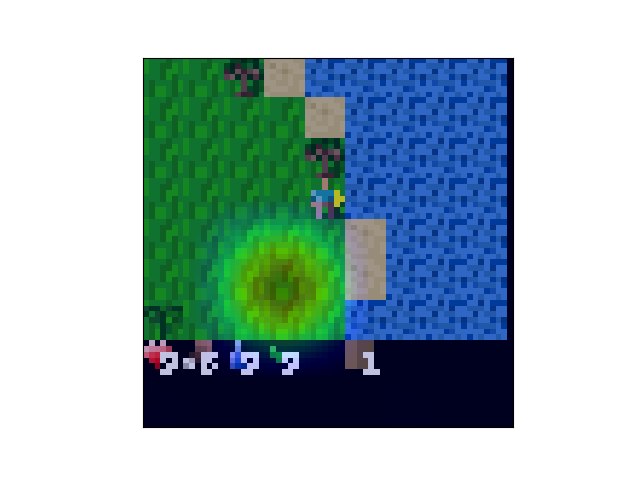} \\ 
\bottomrule 
\end{tabular}
\label{Tab:heatmaps}
\end{table} 

\begin{table}\sffamily
\centering
\caption{Example crops from codes.}
\begin{tabular}{l*8{c@{\hspace{4mm}}}}
\toprule
Code 46 & 
\includegraphics[align=c, width=0.08\textwidth]{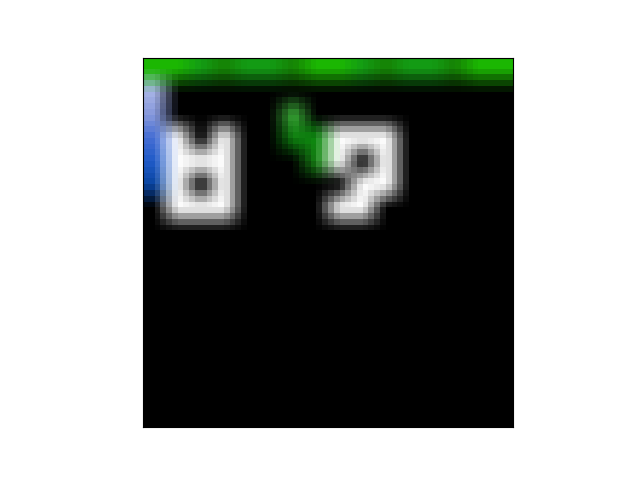} & \includegraphics[align=c, width=0.08\textwidth]{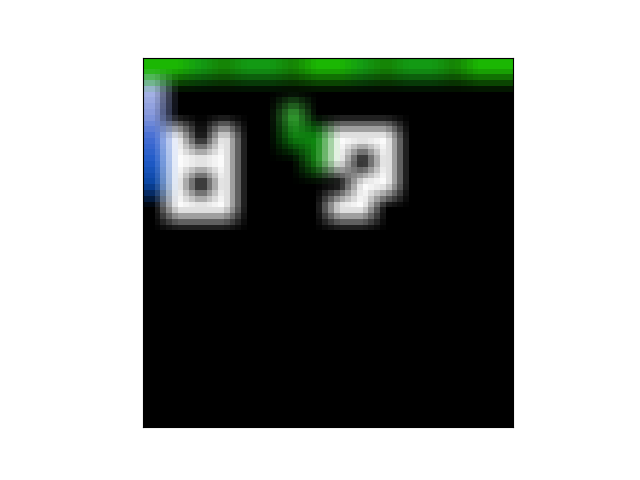} & \includegraphics[align=c, width=0.08\textwidth]{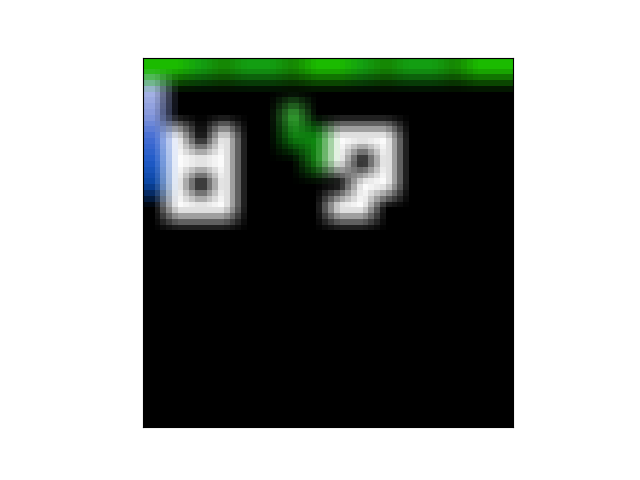} & \includegraphics[align=c, width=0.08\textwidth]{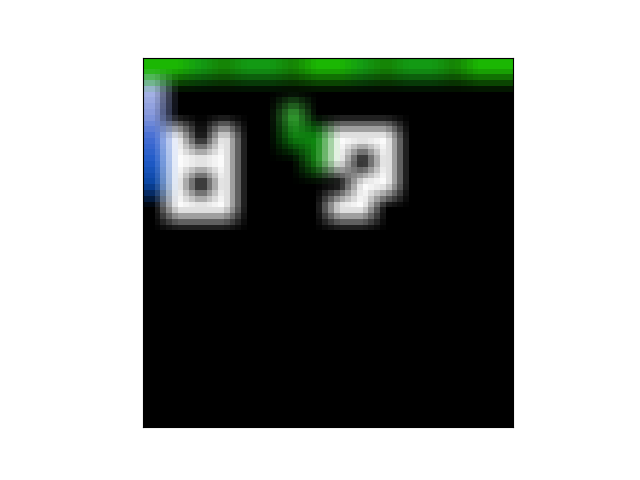} & \includegraphics[align=c, width=0.08\textwidth]{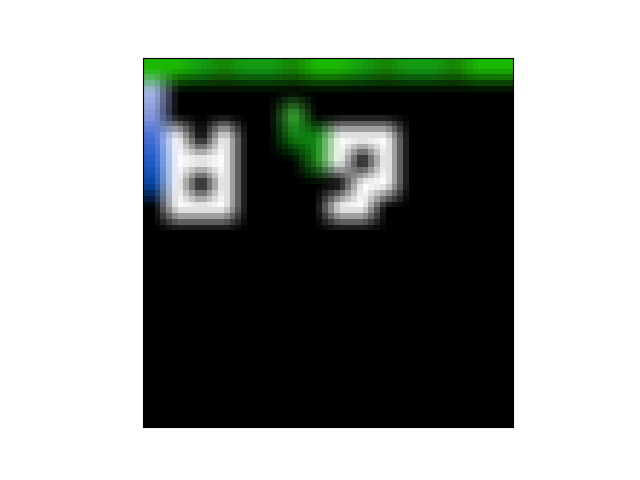} & \includegraphics[align=c, width=0.08\textwidth]{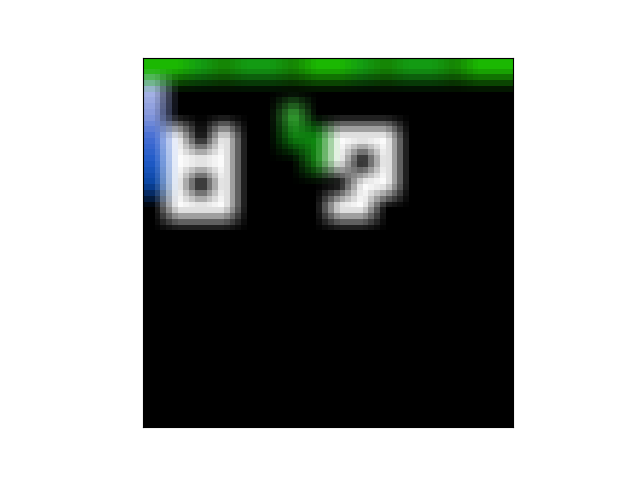} & \includegraphics[align=c, width=0.08\textwidth]{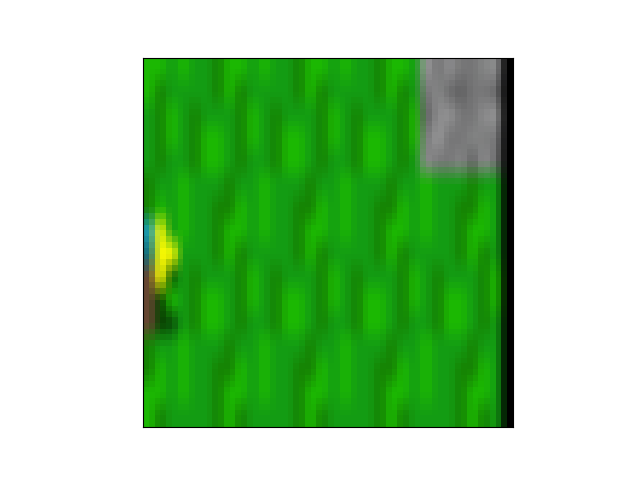} & \includegraphics[align=c, width=0.08\textwidth]{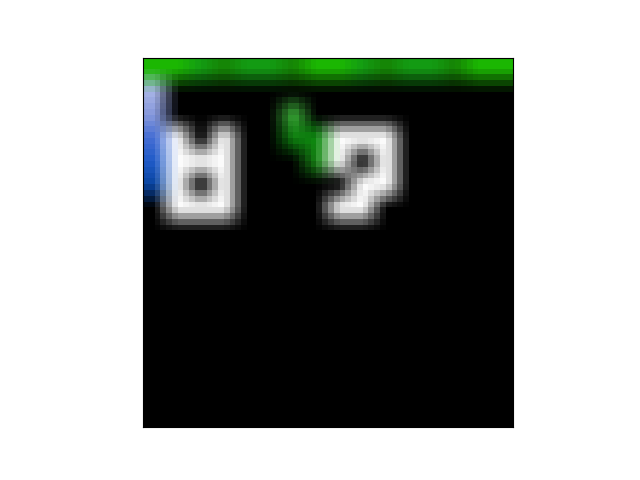} \\ 
Code 162 & 
\includegraphics[align=c, width=0.08\textwidth]{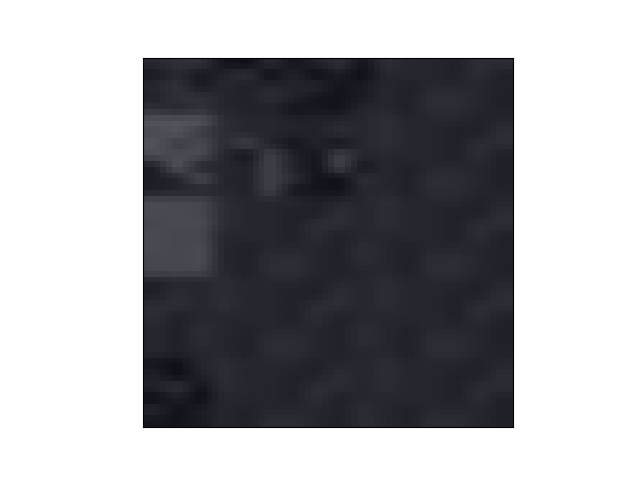} & \includegraphics[align=c, width=0.08\textwidth]{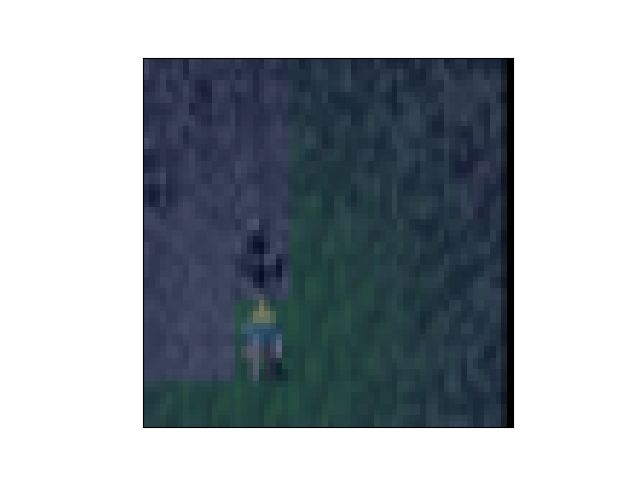} & \includegraphics[align=c, width=0.08\textwidth]{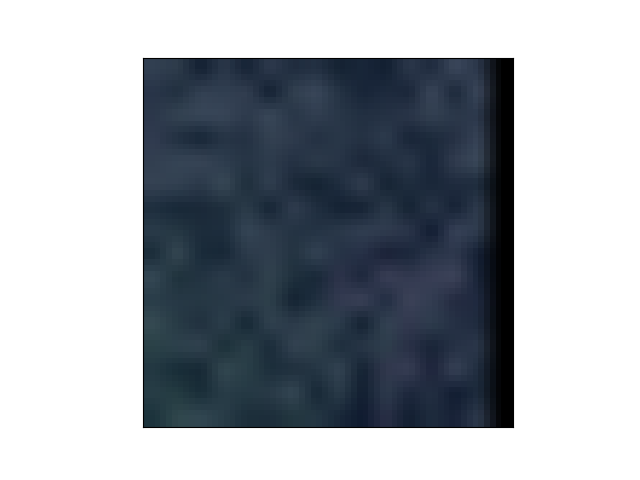} & \includegraphics[align=c, width=0.08\textwidth]{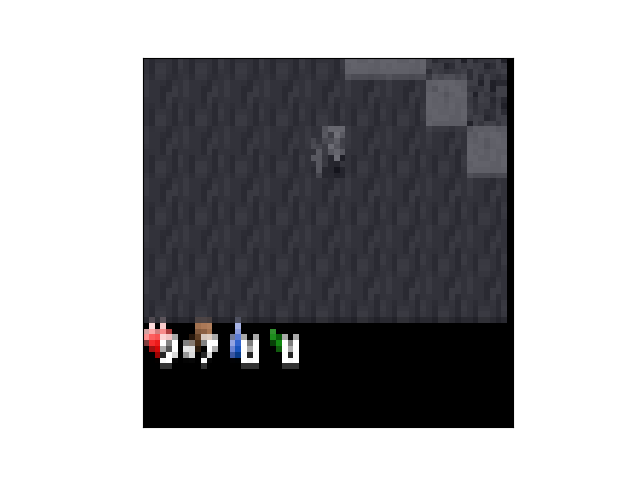} & \includegraphics[align=c, width=0.08\textwidth]{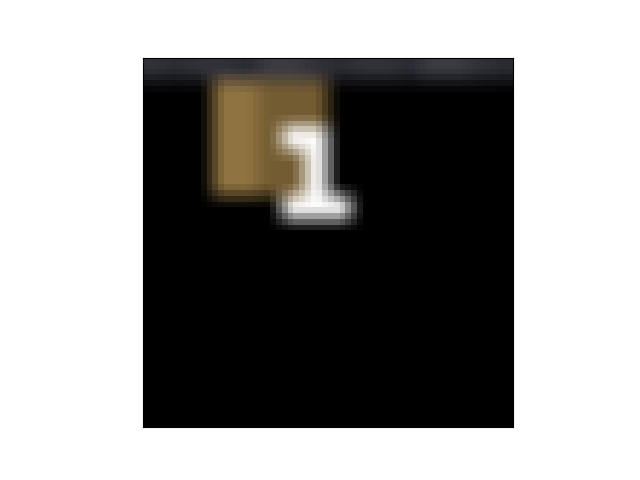} & \includegraphics[align=c, width=0.08\textwidth]{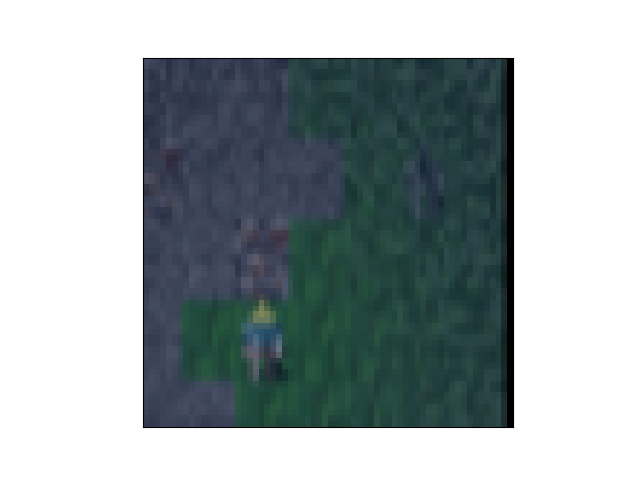} & \includegraphics[align=c, width=0.08\textwidth]{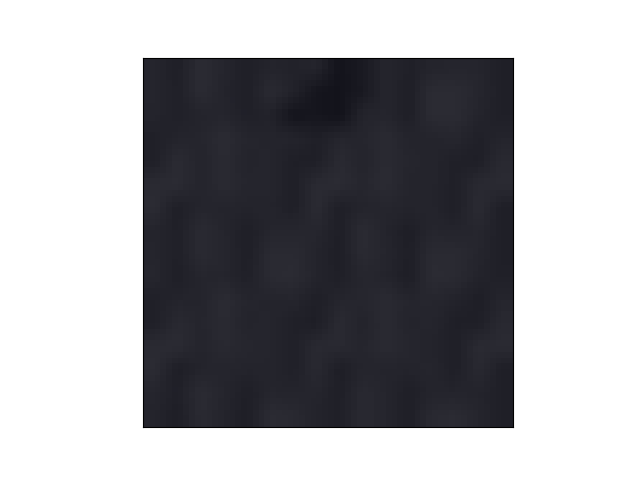} & \includegraphics[align=c, width=0.08\textwidth]{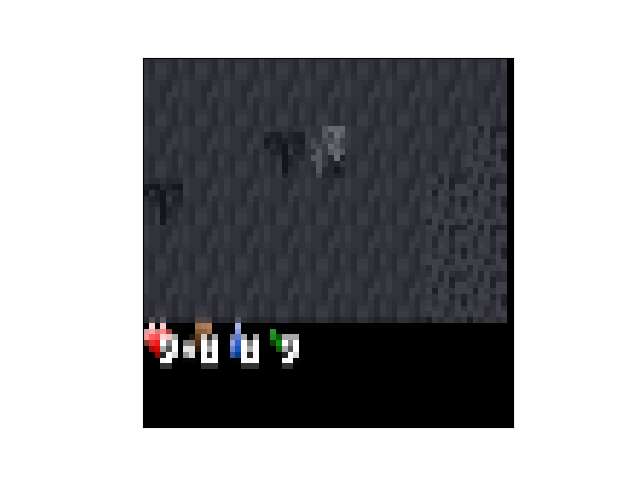} \\ 
Code 498 & 
\includegraphics[align=c, width=0.08\textwidth]{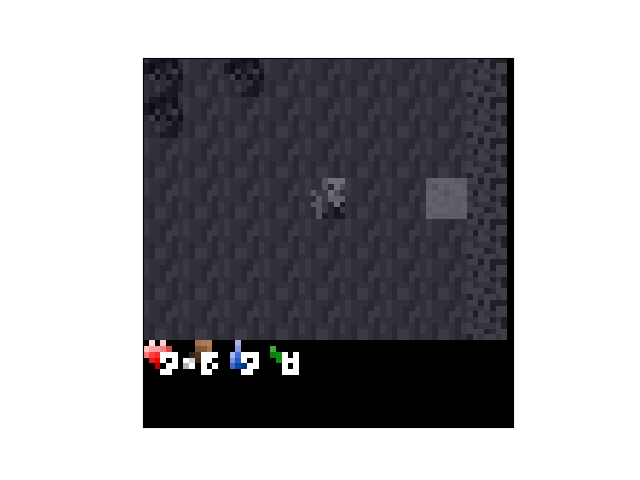} & \includegraphics[align=c, width=0.08\textwidth]{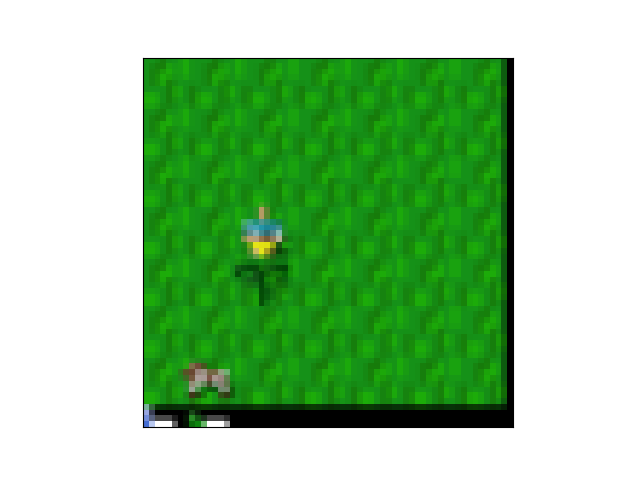} & \includegraphics[align=c, width=0.08\textwidth]{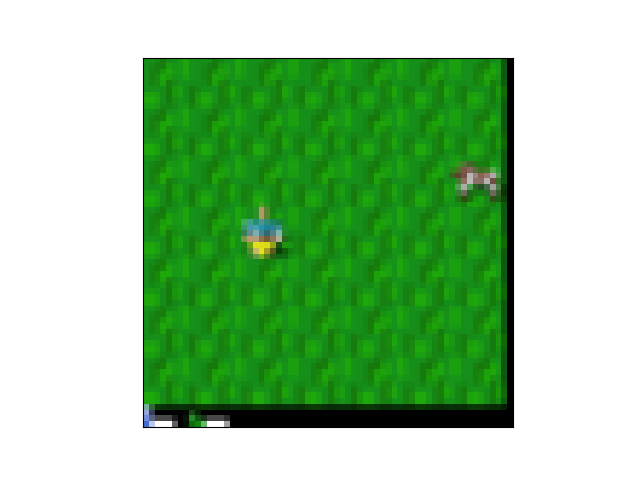} & \includegraphics[align=c, width=0.08\textwidth]{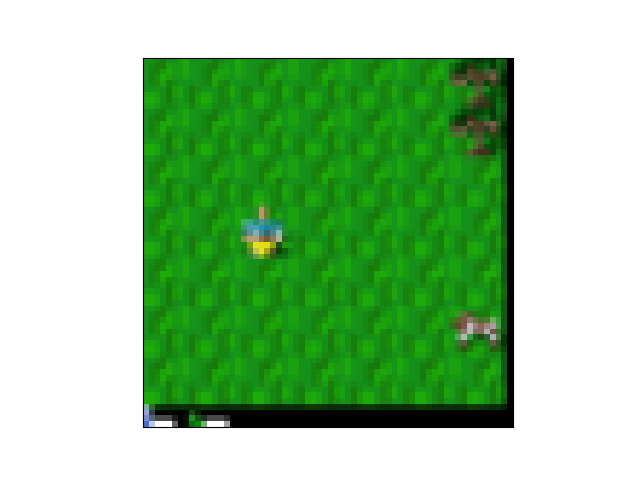} & \includegraphics[align=c, width=0.08\textwidth]{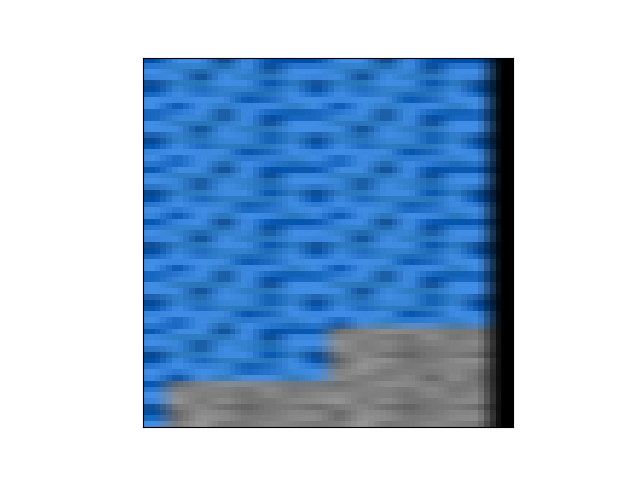} & \includegraphics[align=c, width=0.08\textwidth]{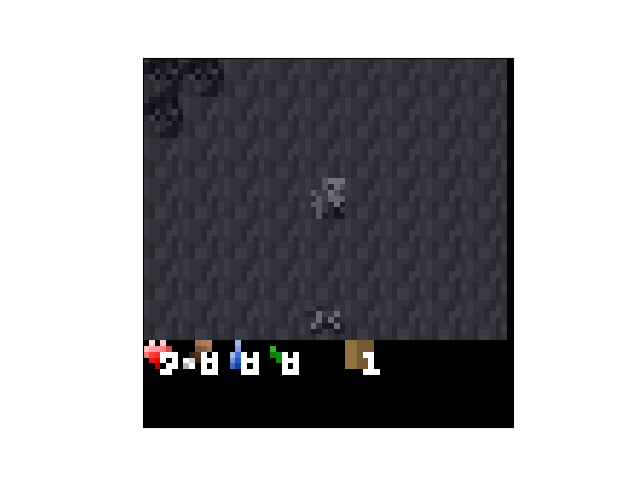} & \includegraphics[align=c, width=0.08\textwidth]{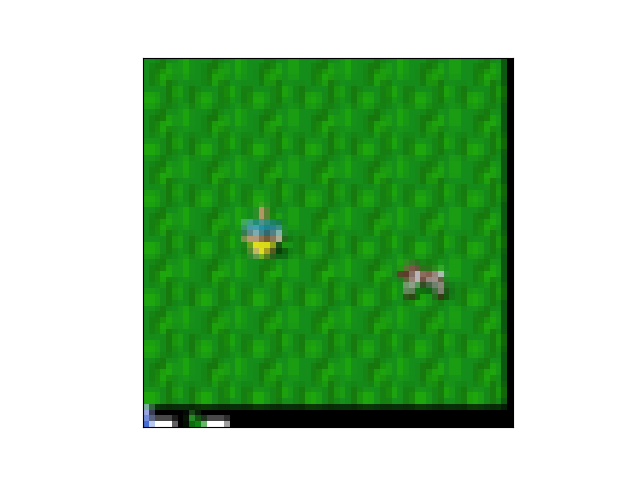} & \includegraphics[align=c, width=0.08\textwidth]{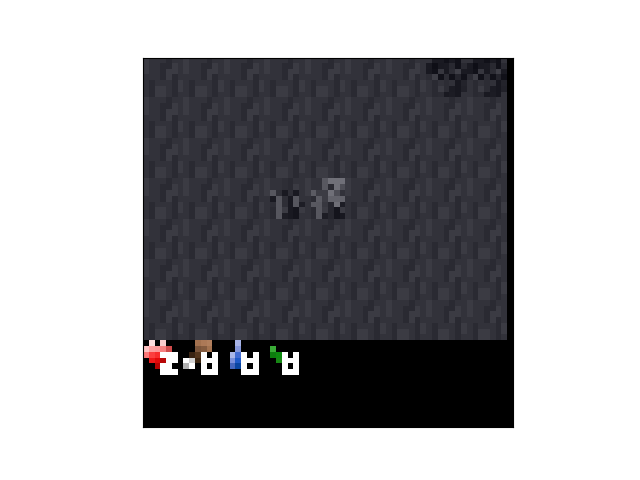} \\ 
\bottomrule 
\end{tabular}
\label{Tab:crops}
\end{table}

To quantify code interpretability, we compared the embeddings for each crop from the last hidden layer of a pre-trained, frozen ResNet50~\citep{he2016deep}  model trained on ImageNet-1k~\citep{russakovsky2015imagenet}.
Given these crop embeddings, we calculated the mean embedding of each code and measured code consistency as the average cosine similarity between the mean code embedding and all embeddings for that code.

\begin{figure*}[t!]
    \centering
    \begin{subfigure}[t]{0.5\textwidth}
        \centering        \includegraphics[width=1\textwidth]{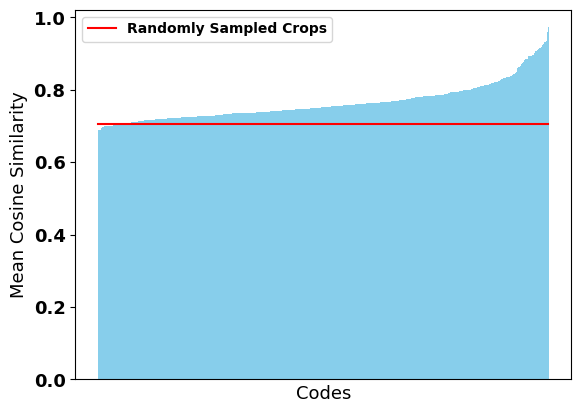}
        \caption{Mean cosine similarity of cropped images}
        \label{fig:cos-sim}
    \end{subfigure}%
    ~ 
    \begin{subfigure}[t]{0.5\textwidth}
        \centering        \includegraphics[width=1\textwidth]{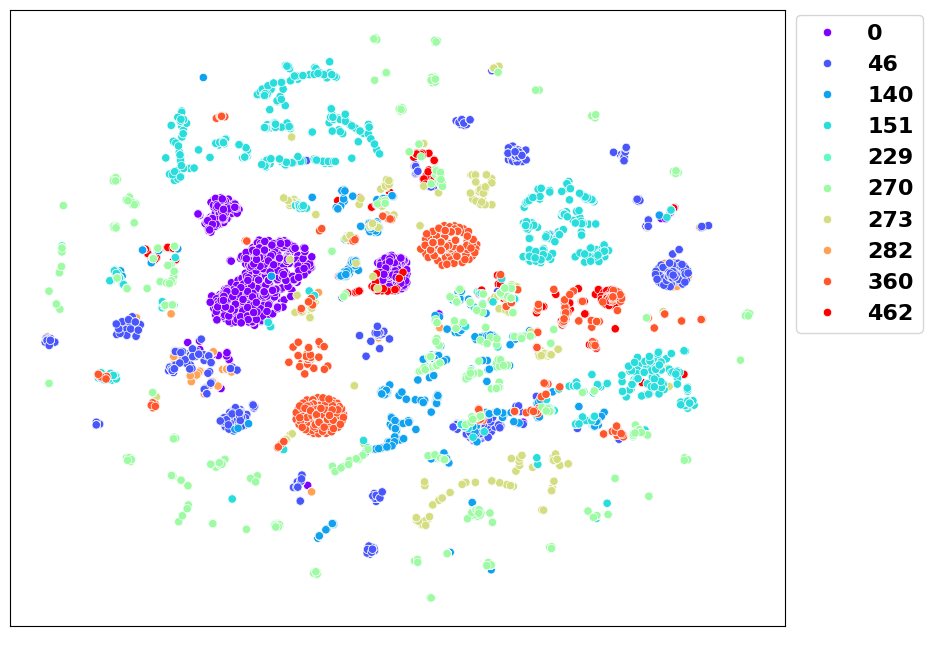}
        \caption{t-SNE of codes with high cosine similarity}
        \label{fig:tsne}
    \end{subfigure}
    \caption{The data from our GradCAM experiments show that VQ codes are inconsistent quantitatively (low mean cosine similarity) and qualitatively (in the lack of t-SNE separation).}
    \label{fig:code-var}
\end{figure*}

In Figure \ref{fig:cos-sim} we show the mean cosine similarity results. The similarity rises gradually with a spike for the best codes.
As a baseline, we calculate the mean cosine similarity of randomly sampled crops, averaged over ten trials, shown as the red line in the plot.
The minimal difference between the random crops and the majority of individual codes suggests at best there are a handful of consistent codes.
In Figure \ref{fig:tsne}, this is examined further with a t-SNE plot of the embeddings for the ten codes with the highest mean cosine similarity and at least 500 embeddings, to ensure similar order of magnitude of samples.
We observe only a handful of these codes with strong clustering.
In addition, some codes have multiple clusters, which makes a code's semantic meaning less clear. 

\section{Discussion}
As reported above, the model learned a limited number of codes that capture entities such as grass, water, and resources in the inventory.
In Appendix \ref{sec:appendix3},  we examine co-occurring codes to find additional interpretable examples occurring in superposition~\cite{elhage2022superposition}, however, they occur very sparingly.
Outside of these instances, the codes lacked a consistent semantic theme.
Thus, VQ alone cannot provide interpretability of transition models in MBRL.
One of the reasons we believe this happened is that VQ offers insufficient constraints to enforce semantic disentanglement of codes.
Specifically, the model is always trained to reconstruct the entire image using a combination of codes, and due to the depth of the CNN in our model, the latent vectors have all seen the entire image.
This creates a condition where there is no need for the model to cleanly isolate entities in the image to a single code.
Thus, indicating there are characteristics of discretization that matter to ensure semantic associations, a deeper study of which is left for future work.

\bibliography{main}
\bibliographystyle{rlc}

\appendix

\section{Code Frequency}
\label{sec:appendix1}

\begin{figure}[ht!]
    \centering    \includegraphics[width=0.6\textwidth]{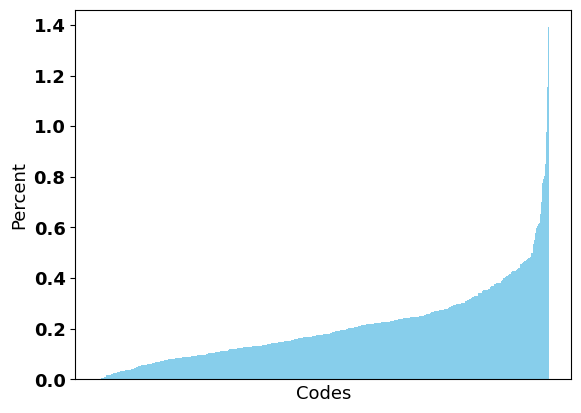}
    \caption{Percentage of each code's occurrence in the dataset.}
    \label{fig:code-usage}
\end{figure}

In order to decide which codes would be the most beneficial to examine for consistency, we organized them by how frequently they occur in the dataset.
As seen in Figure \ref{fig:code-usage}, the model is not overly reliant on any codes because the most frequent one occurs in barely over 1\%.
However, there is a significant dropoff between the top ten codes and the rest, which is the reason we presented heatmaps and crops with three distinct frequencies. The top two codes occur significantly more than the tenth and eleventh codes. The median codes then occur at a much lower rate. We chose to focus on the higher frequency end of the distribution because we believe even if the low frequency codes are consistent, they will not occur enough to provide useful interpretability, since in many observations, none of those codes will appear.

\section{Consistent and Inconsistent Code Crops}
\label{sec:appendix2}

\begin{table}[ht!]\sffamily
\centering
\caption{Additional examples of consistent (top three) and inconsistent (bottom three) codes.}
\begin{tabular}{l*8{c@{\hspace{4mm}}}}
\toprule
Code 0  &
\includegraphics[align=c, width=0.08\textwidth]{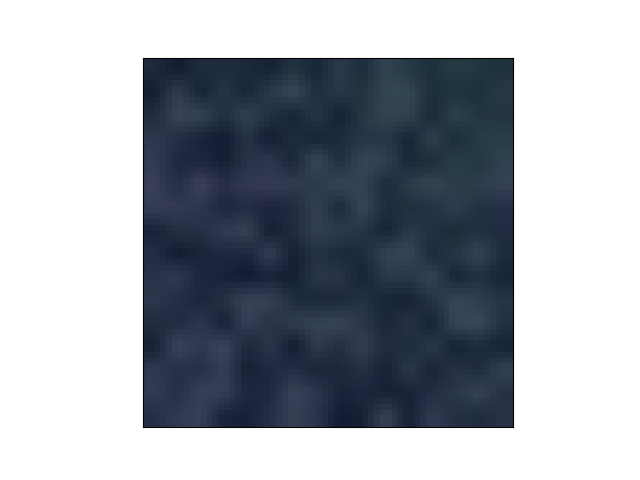} & \includegraphics[align=c, width=0.08\textwidth]{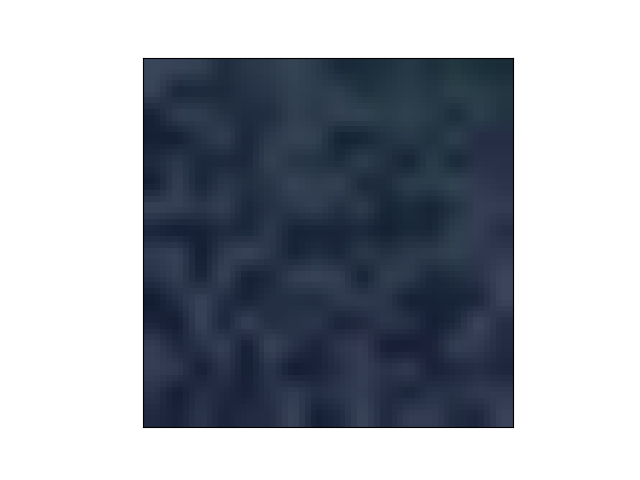} & \includegraphics[align=c, width=0.08\textwidth]{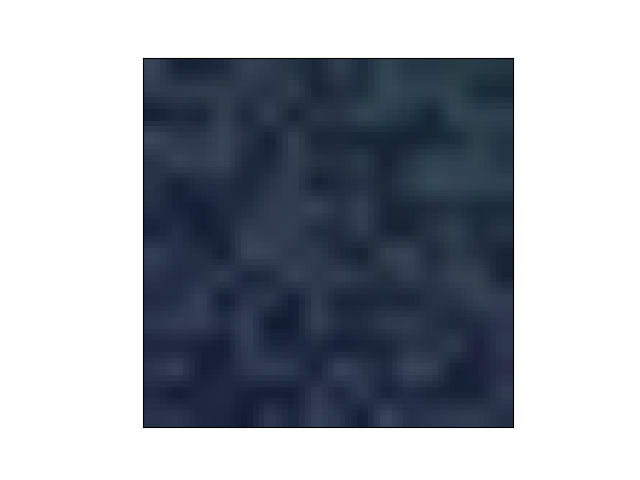} & \includegraphics[align=c, width=0.08\textwidth]{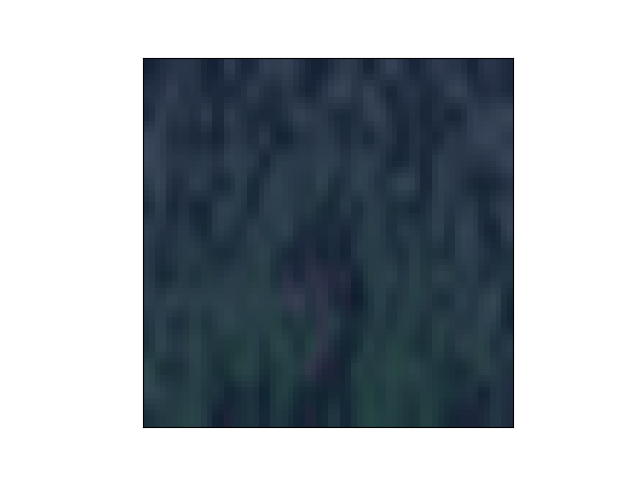} & \includegraphics[align=c, width=0.08\textwidth]{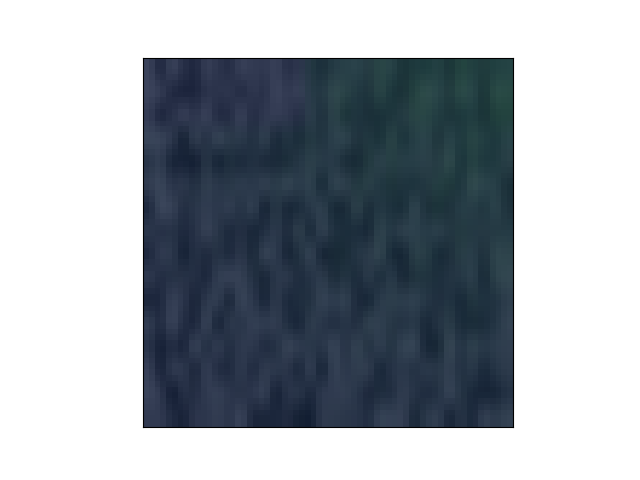} & \includegraphics[align=c, width=0.08\textwidth]{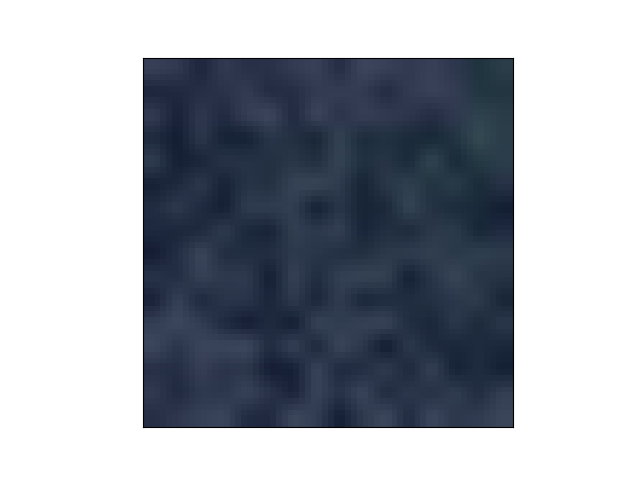} & \includegraphics[align=c, width=0.08\textwidth]{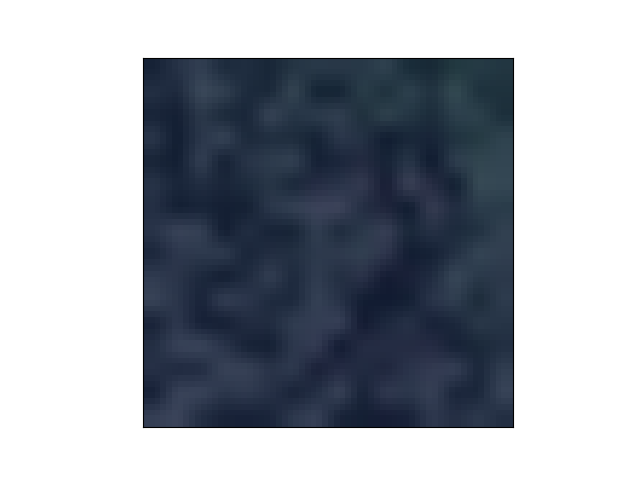} & \includegraphics[align=c, width=0.08\textwidth]{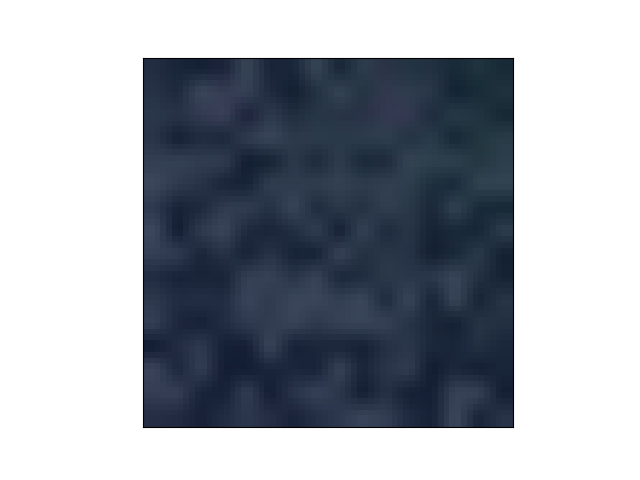} \\ 
Code 360 & 
\includegraphics[align=c, width=0.08\textwidth]{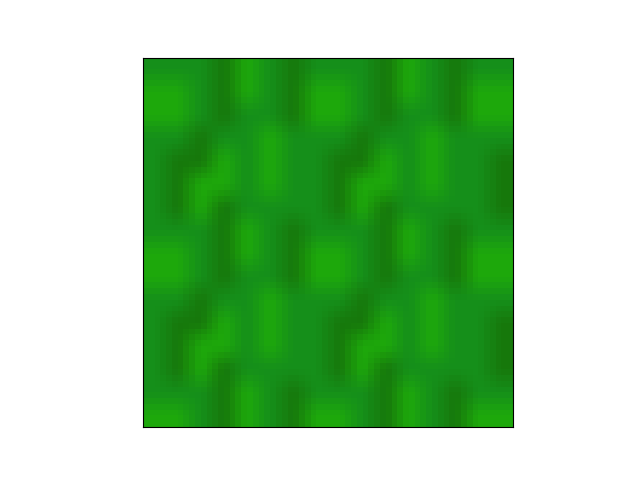} & \includegraphics[align=c, width=0.08\textwidth]{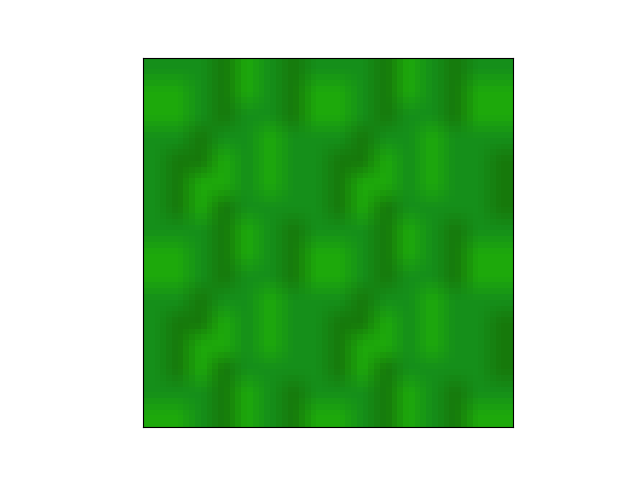} & \includegraphics[align=c, width=0.08\textwidth]{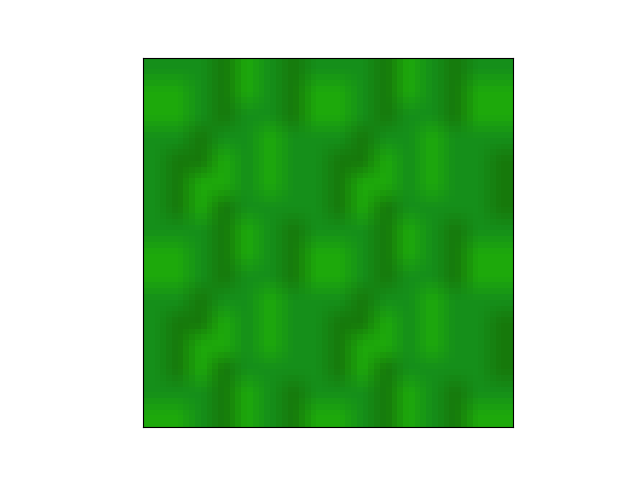} & \includegraphics[align=c, width=0.08\textwidth]{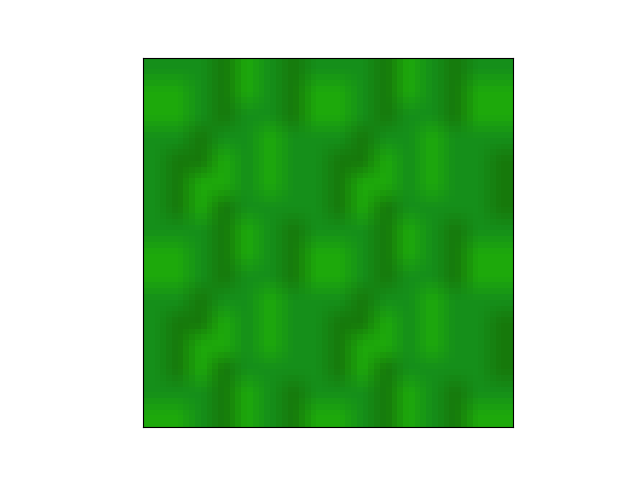} & \includegraphics[align=c, width=0.08\textwidth]{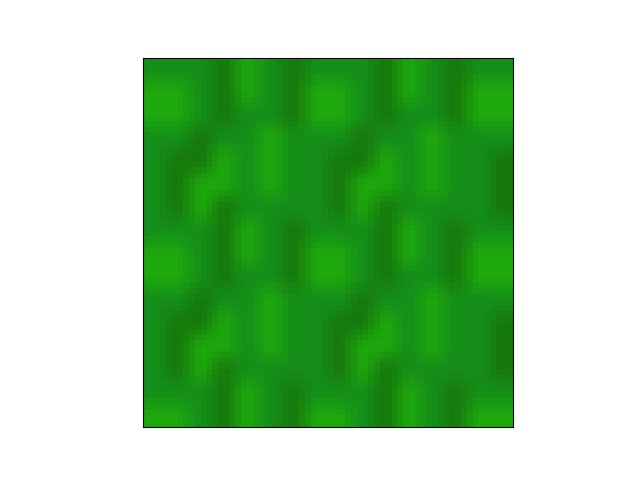} & \includegraphics[align=c, width=0.08\textwidth]{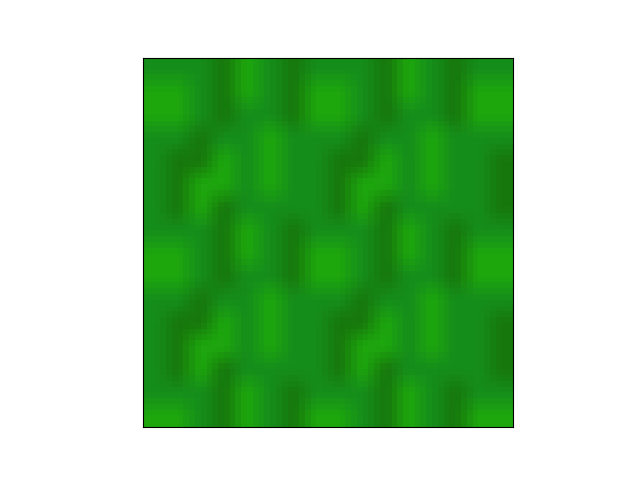} & \includegraphics[align=c, width=0.08\textwidth]{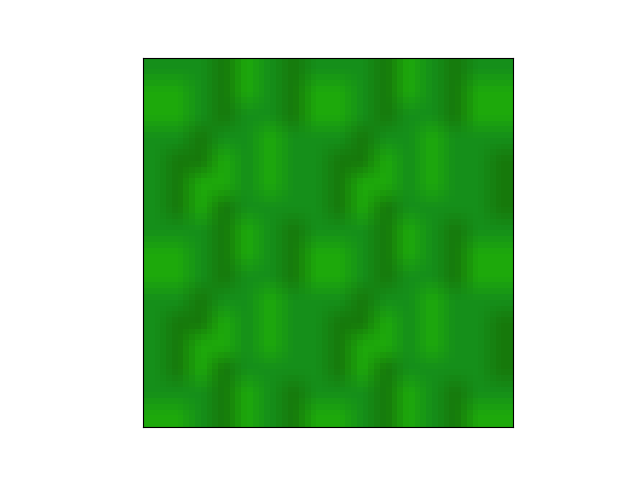} & \includegraphics[align=c, width=0.08\textwidth]{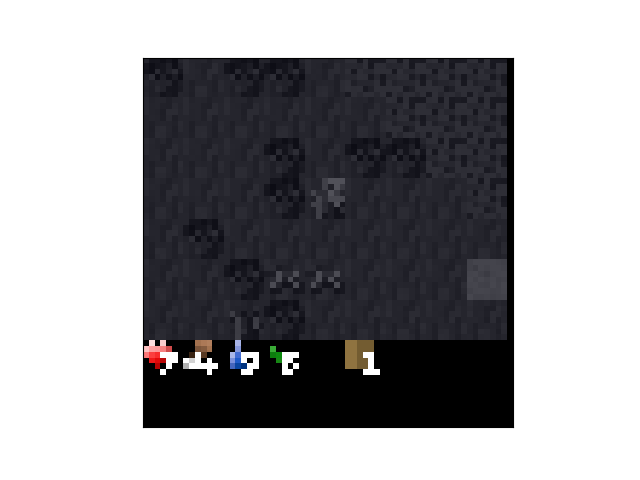} \\ 
Code 192 & 
\includegraphics[align=c, width=0.08\textwidth]{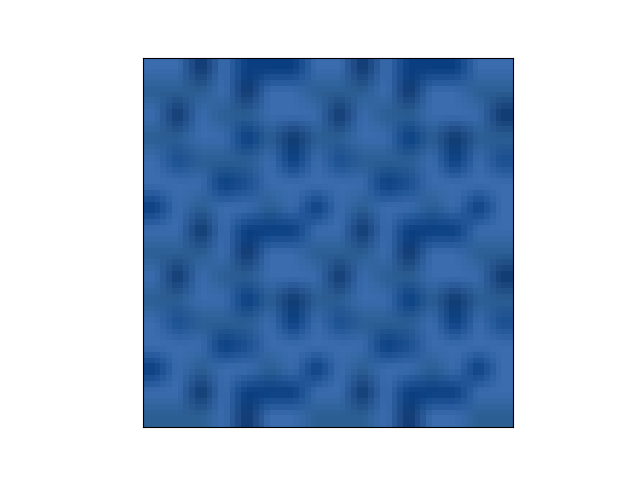} & \includegraphics[align=c, width=0.08\textwidth]{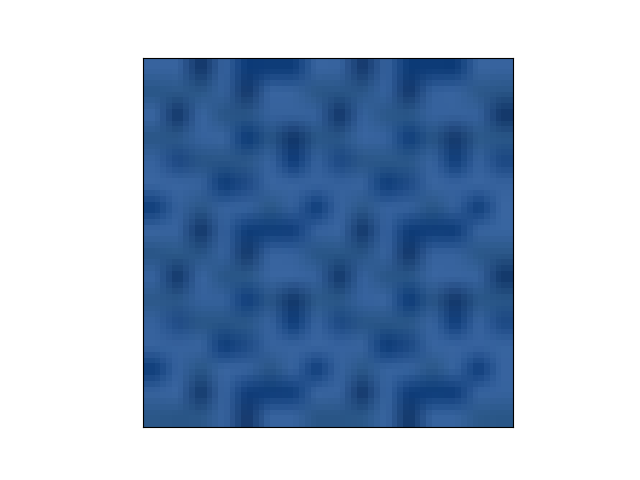} & \includegraphics[align=c, width=0.08\textwidth]{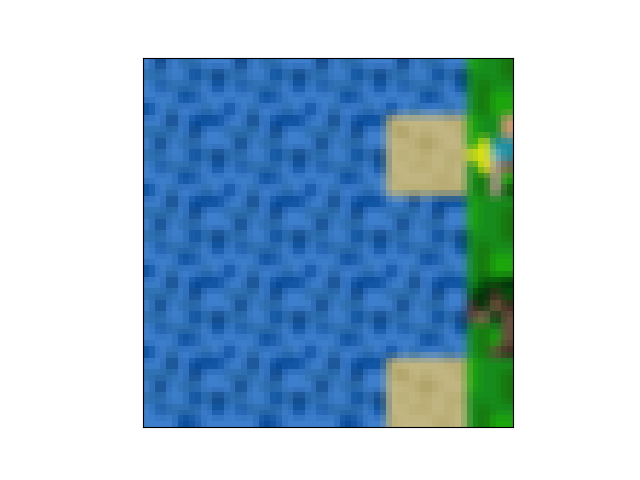} & \includegraphics[align=c, width=0.08\textwidth]{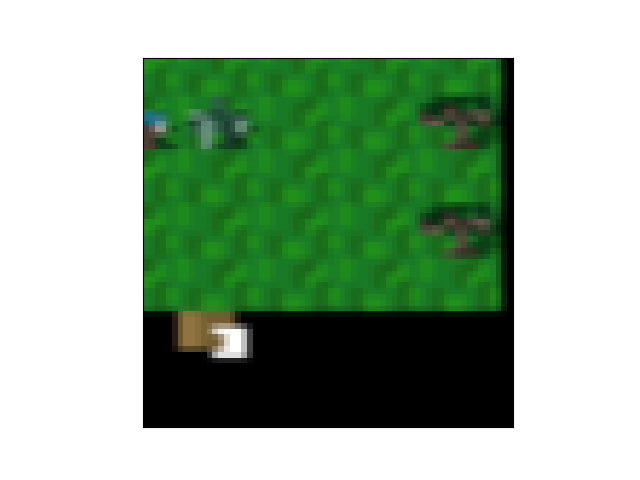} & \includegraphics[align=c, width=0.08\textwidth]{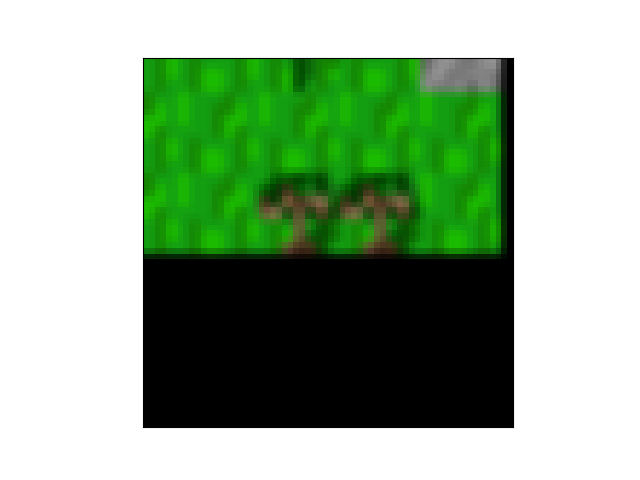} & \includegraphics[align=c, width=0.08\textwidth]{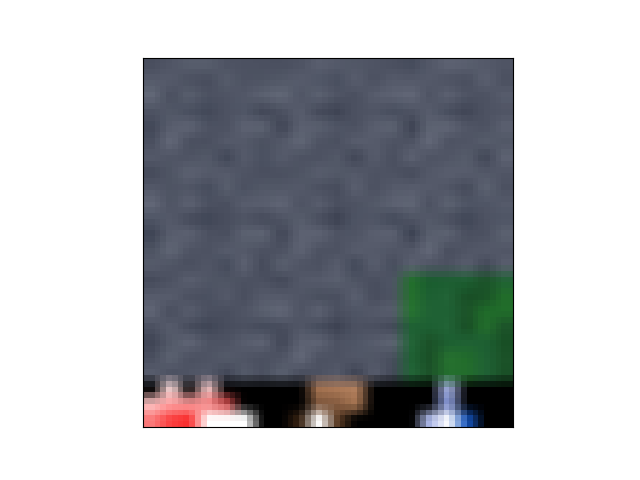} & \includegraphics[align=c, width=0.08\textwidth]{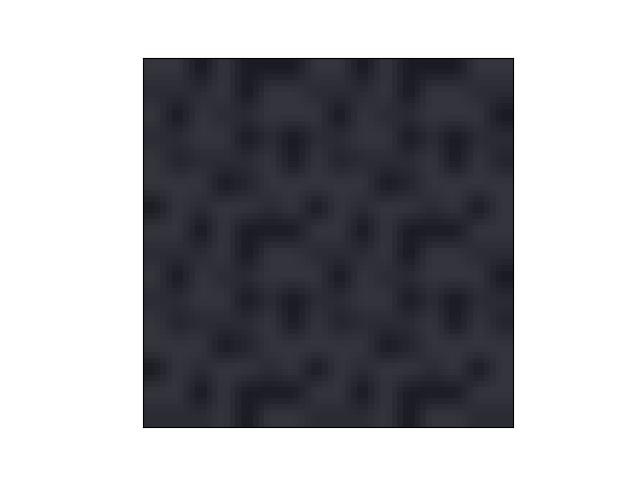} & \includegraphics[align=c, width=0.08\textwidth]{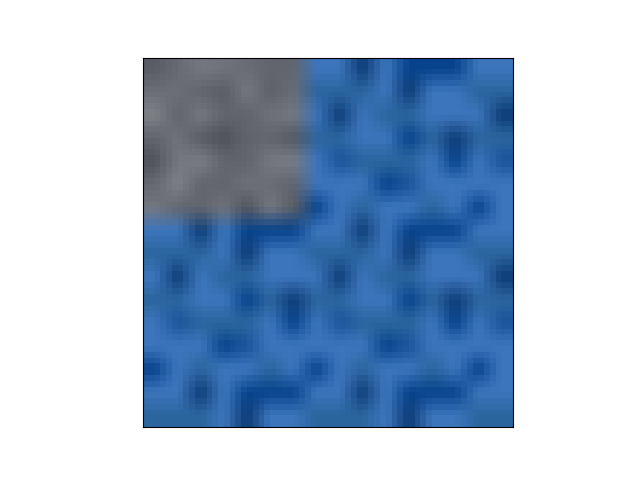} \\
\midrule
Code 15 & 
\includegraphics[align=c, width=0.08\textwidth]{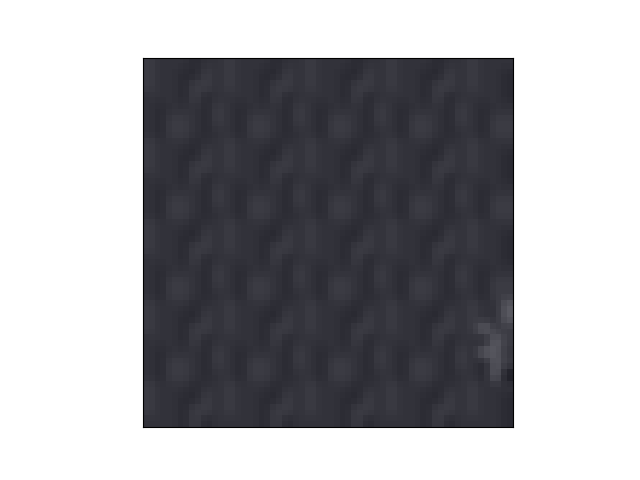} & \includegraphics[align=c, width=0.08\textwidth]{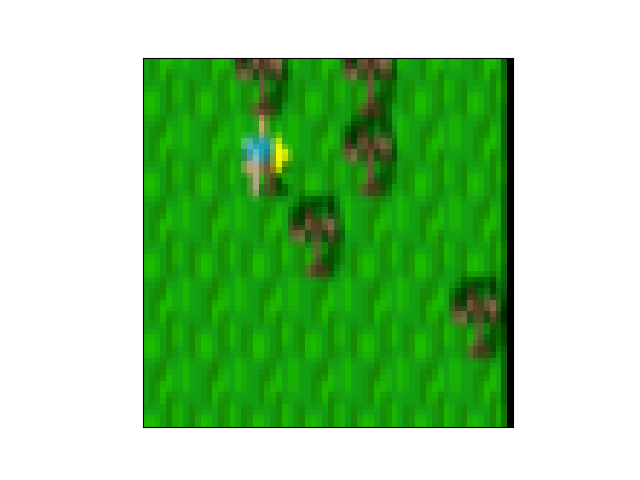} & \includegraphics[align=c, width=0.08\textwidth]{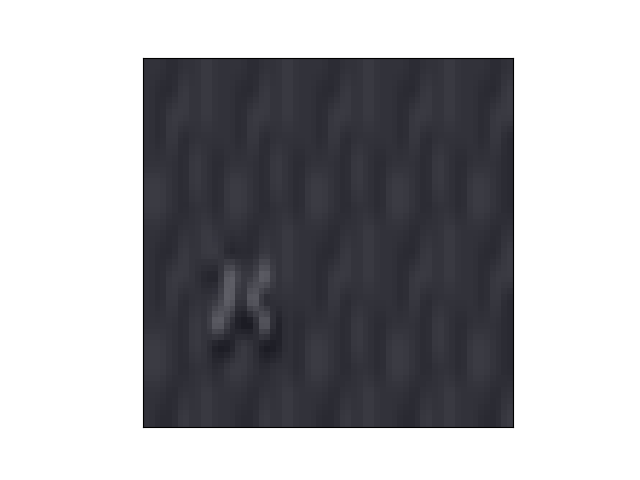} & \includegraphics[align=c, width=0.08\textwidth]{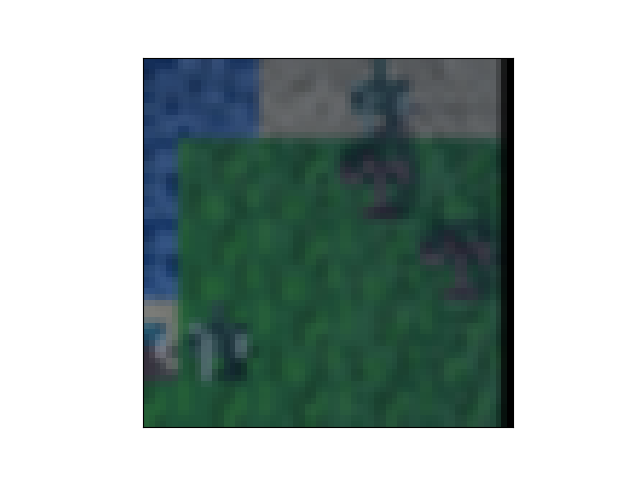} & \includegraphics[align=c, width=0.08\textwidth]{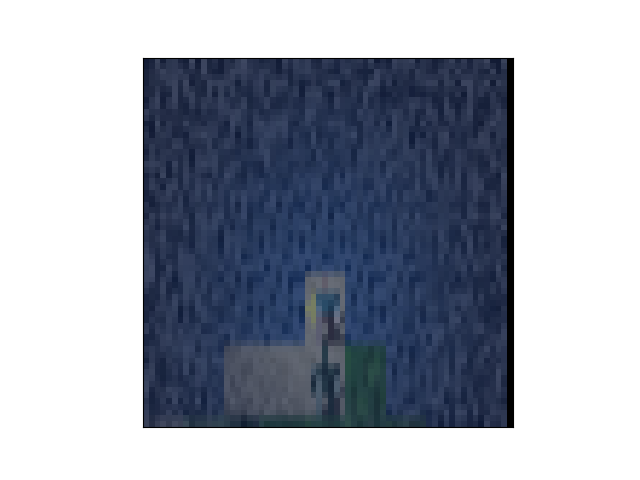} & \includegraphics[align=c, width=0.08\textwidth]{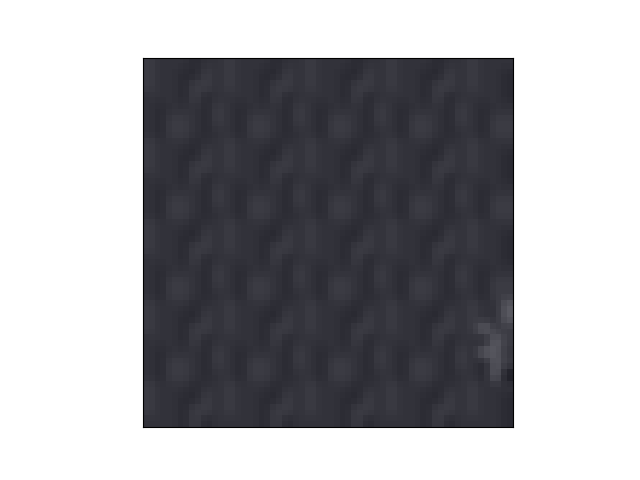} & \includegraphics[align=c, width=0.08\textwidth]{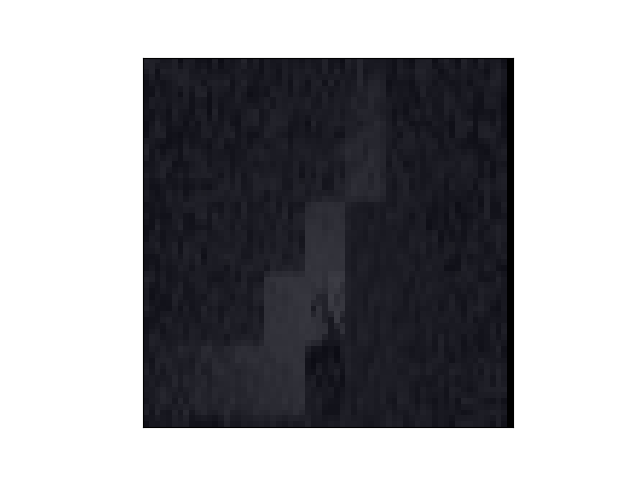} & \includegraphics[align=c, width=0.08\textwidth]{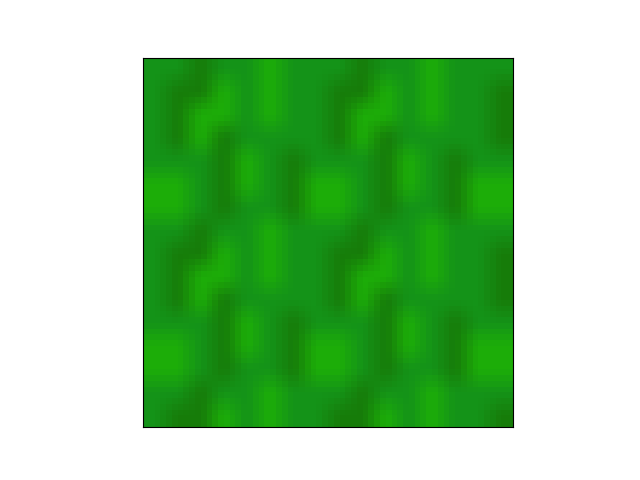} \\ 
Code 329 & 
\includegraphics[align=c, width=0.08\textwidth]{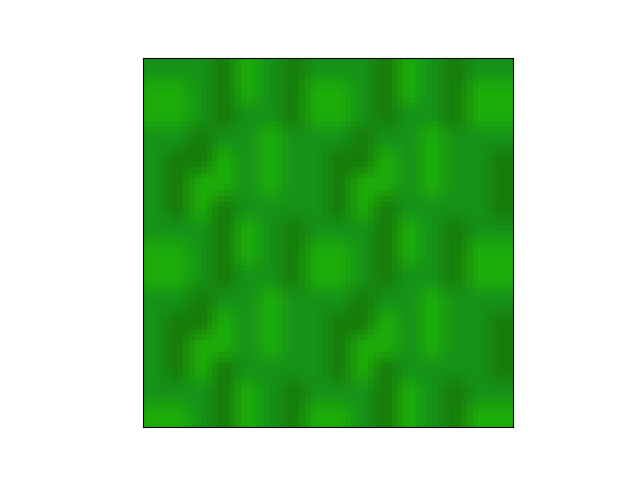} & \includegraphics[align=c, width=0.08\textwidth]{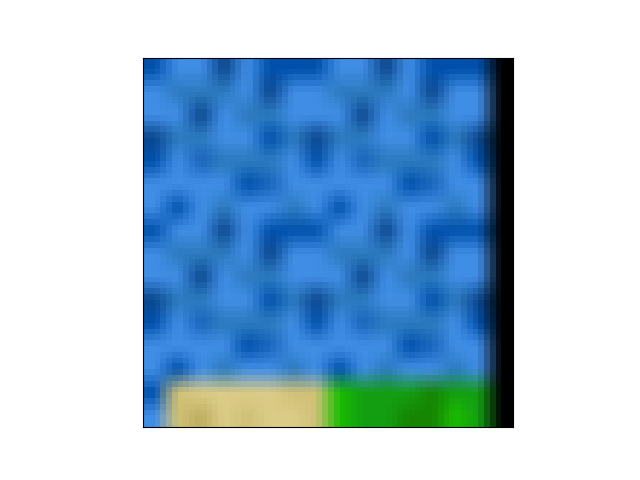} & \includegraphics[align=c, width=0.08\textwidth]{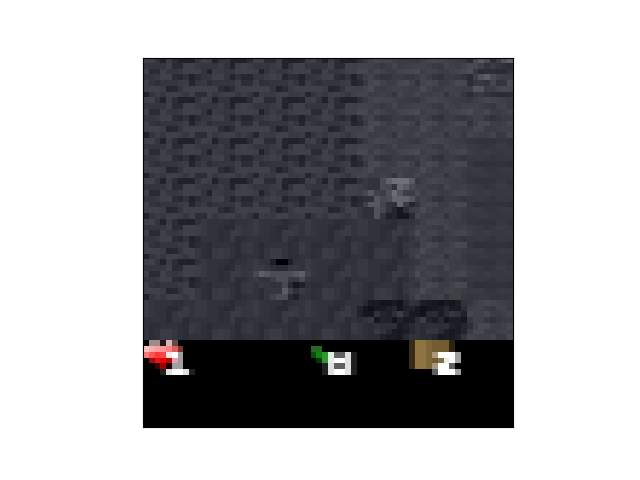} & \includegraphics[align=c, width=0.08\textwidth]{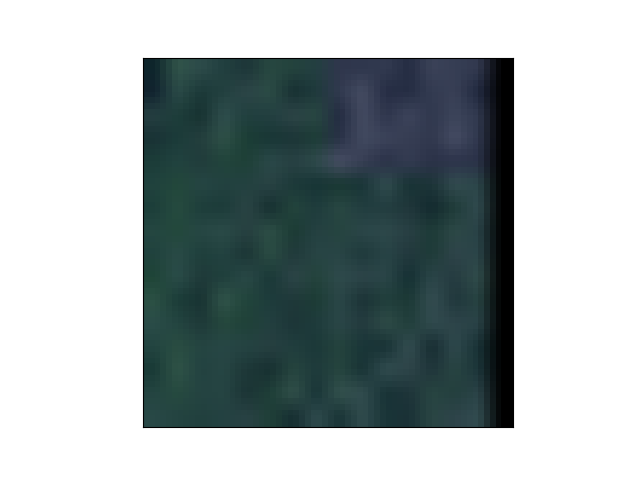} & \includegraphics[align=c, width=0.08\textwidth]{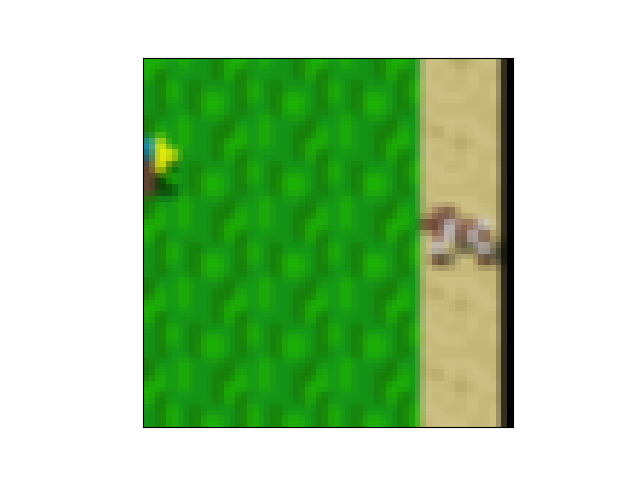} & \includegraphics[align=c, width=0.08\textwidth]{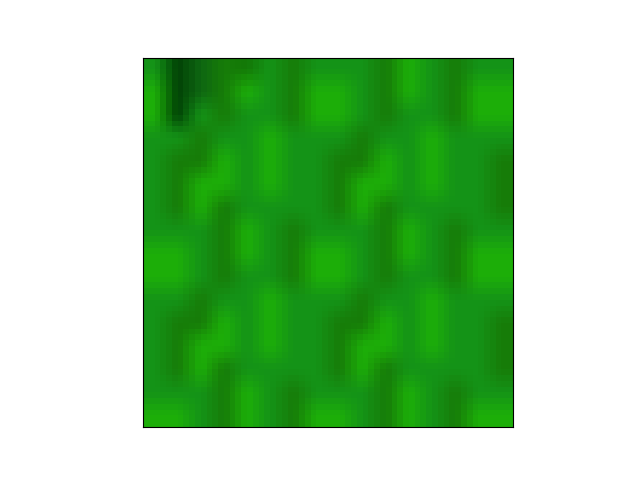} & \includegraphics[align=c, width=0.08\textwidth]{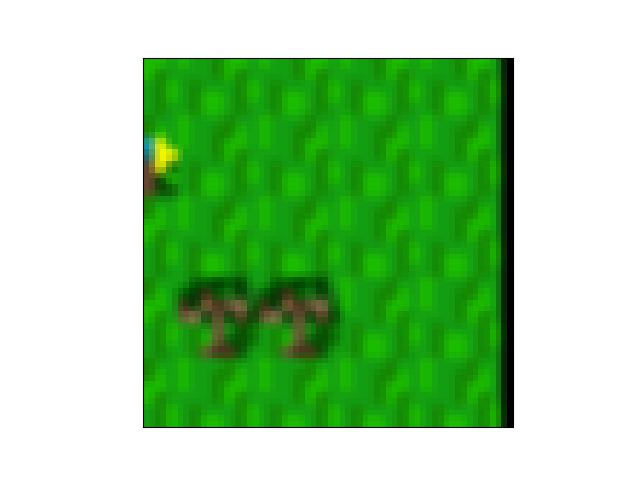} & \includegraphics[align=c, width=0.08\textwidth]{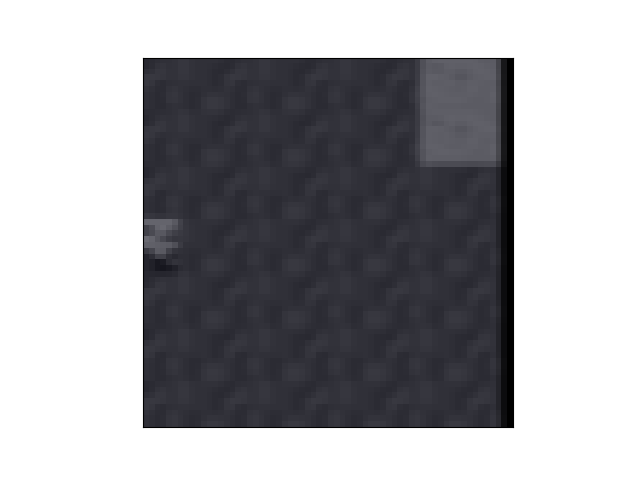} \\ 
Code 438 & 
\includegraphics[align=c, width=0.08\textwidth]{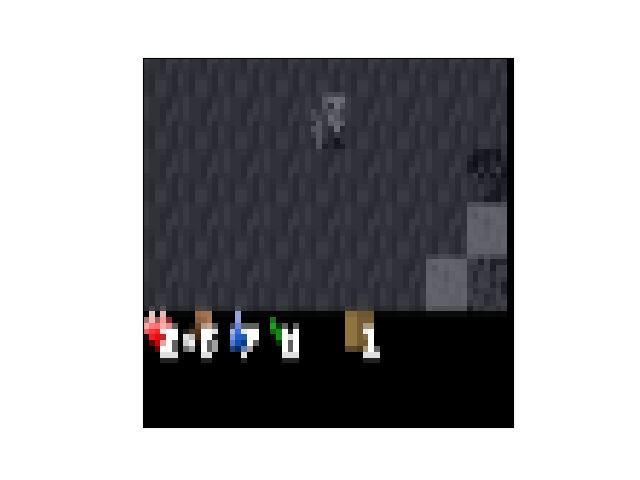} & \includegraphics[align=c, width=0.08\textwidth]{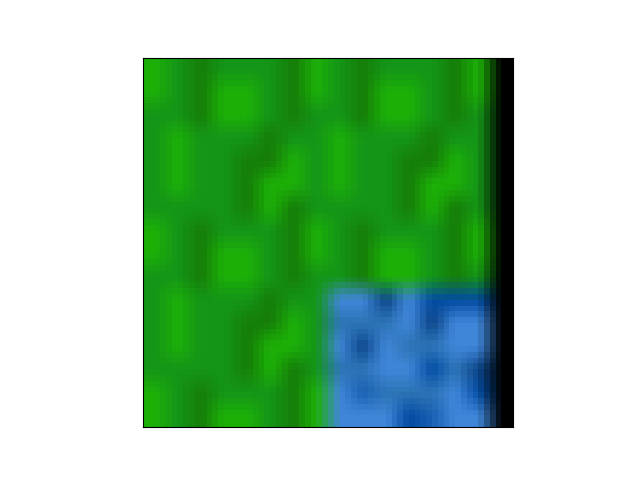} & \includegraphics[align=c, width=0.08\textwidth]{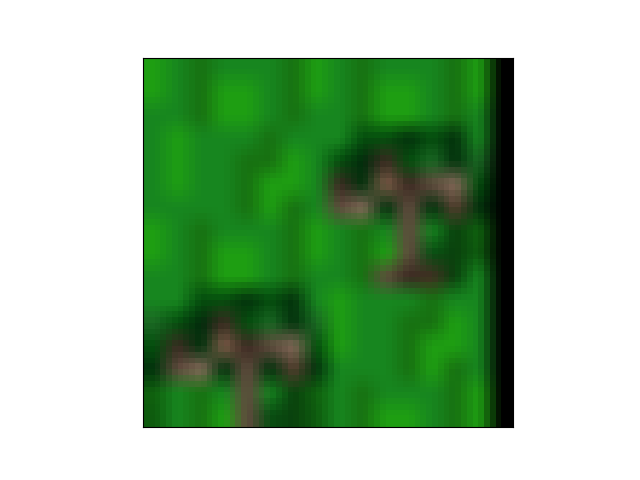} & \includegraphics[align=c, width=0.08\textwidth]{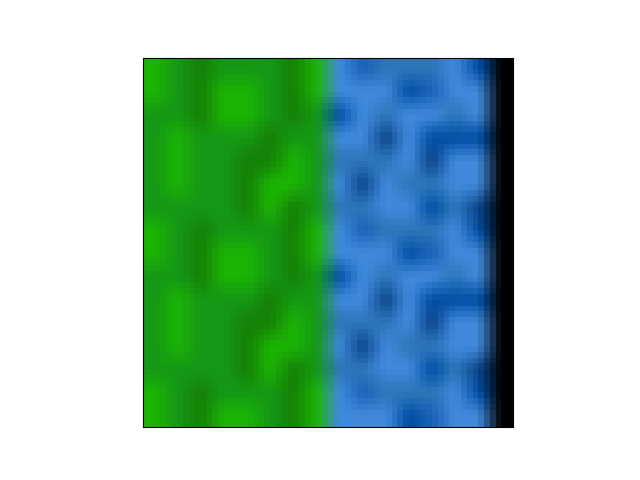} & \includegraphics[align=c, width=0.08\textwidth]{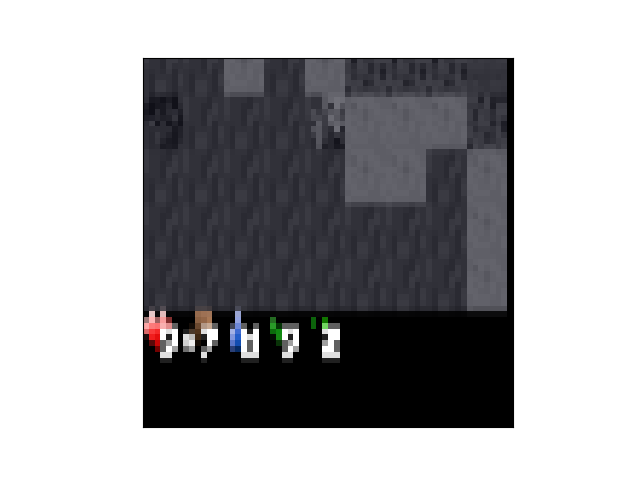} & \includegraphics[align=c, width=0.08\textwidth]{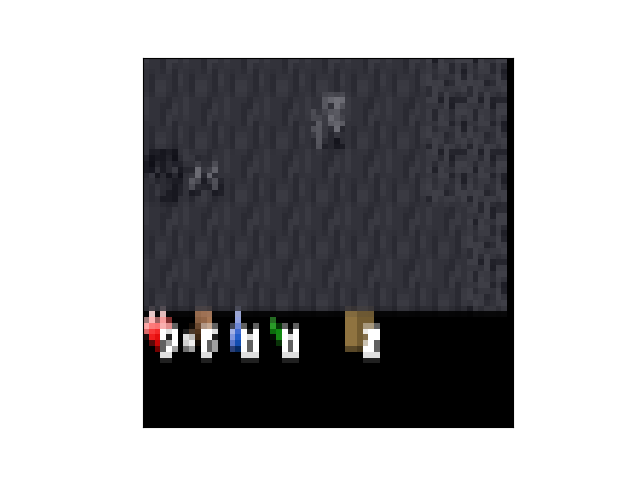} & \includegraphics[align=c, width=0.08\textwidth]{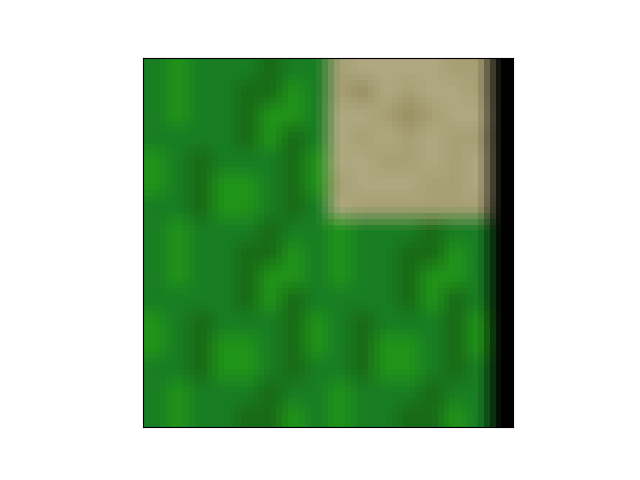} & \includegraphics[align=c, width=0.08\textwidth]{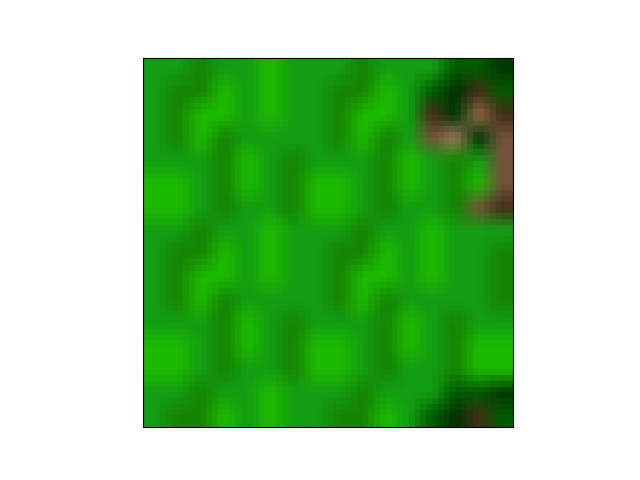} \\ 
\bottomrule 
\end{tabular}
\label{Tab:crops2}
\end{table} 

In Table \ref{Tab:crops2}, we show additional cropped images randomly sampled from six codes.
The top three shown are examples of the more consistent codes we observed.
The first one focuses on stone, while the second captures grass, and the third focuses on water.
Throughout our entire dataset of 477 codes, only a handful are close to as consistent as these.
Even the ones that are have inconsistent instances, like the fourth and fifth crops for code 192, that limit how confidently they can be interpreted.
Below them, we show examples for three inconsistent codes.
The majority of the codes look closer to these when randomly sampling their cropped images.
Across all three of the codes, the crops are widely varied in what they contain, providing no value for interpretability.

\section{Code Co-occurrence}
\label{sec:appendix3}

\begin{figure}[ht!]
    \centering
    \includegraphics[width=0.6\textwidth]{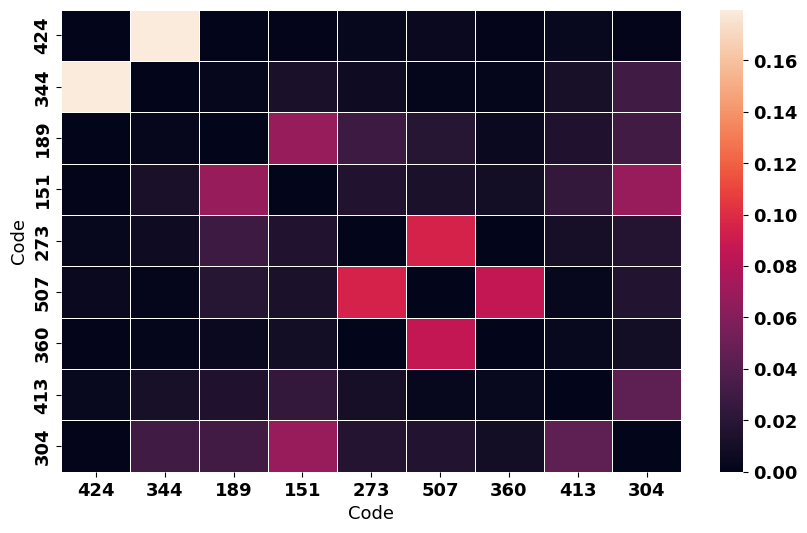}
    \caption{Co-occurrence rate for the ten most frequently co-occurring pairs of codes}
    \label{fig:co-occur}
\end{figure}

One of the potential reasons for the lack of interpretability of codes is that the model learned a combination of codes to capture entities.
In order to investigate this, we found the codes that most frequently appear together and calculated their co-occurrence rate. We define co-occurrence rate as the number of appearances together divided by the average number of times the two codes are used. 
This is related to the mechanistic interpretability concept of superposition~\cite{elhage2022superposition}, where  given a latent feature space represented by a set of $d$-dimensional vectors $v$, optimization tries to use these vectors  to represent more features than they have the dimensional capacity. 
This leads to phenomena such as representing concepts with groups of features, or individual features alternately representing more than one concept depending on the input, both of which we observe in our work~\cite{elhage2022superposition,nanda_2022}.
Figure \ref{fig:co-occur} shows a heatmap of the rates for the ten most frequently co-occurring pairs.
Note the co-occurrence of codes to themselves has been set to zero to better distinguish the rates between different codes.
The highest rate observed was between codes 344 and 424 at around 18\%.

\begin{table}[t!]\sffamily
\centering
\caption{Example crops from co-occurring codes, randomly sampled from observations where they co-occur.}
\begin{tabular}{l*8{c@{\hspace{4mm}}}}
\toprule
Code 424 & 
\includegraphics[align=c, width=0.08\textwidth]{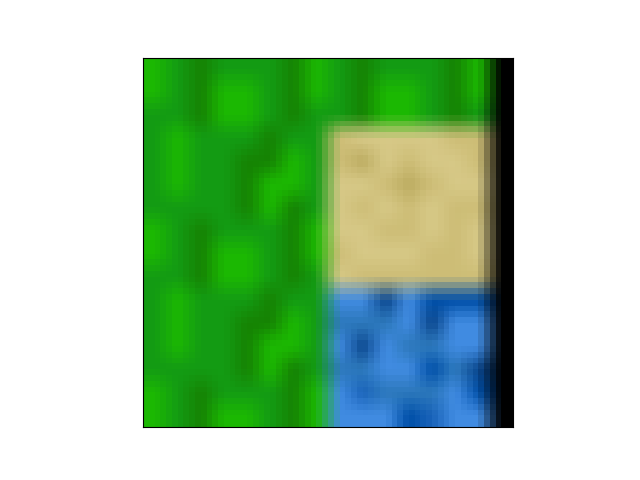} & \includegraphics[align=c, width=0.08\textwidth]{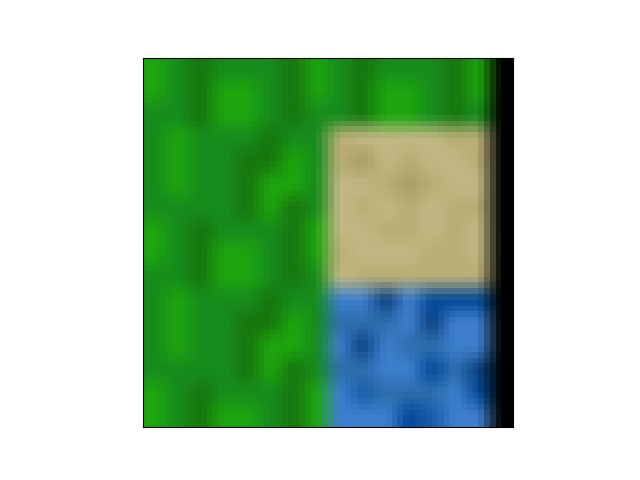} & \includegraphics[align=c, width=0.08\textwidth]{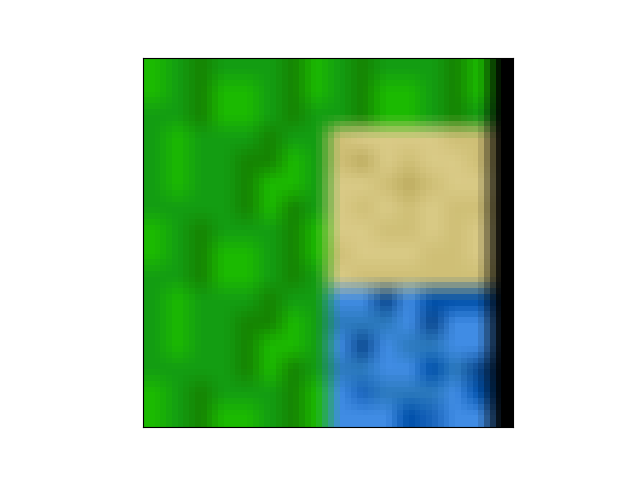} & \includegraphics[align=c, width=0.08\textwidth]{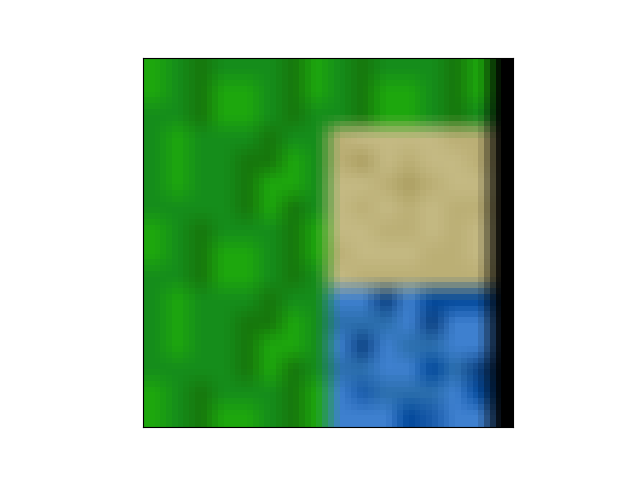} & \includegraphics[align=c, width=0.08\textwidth]{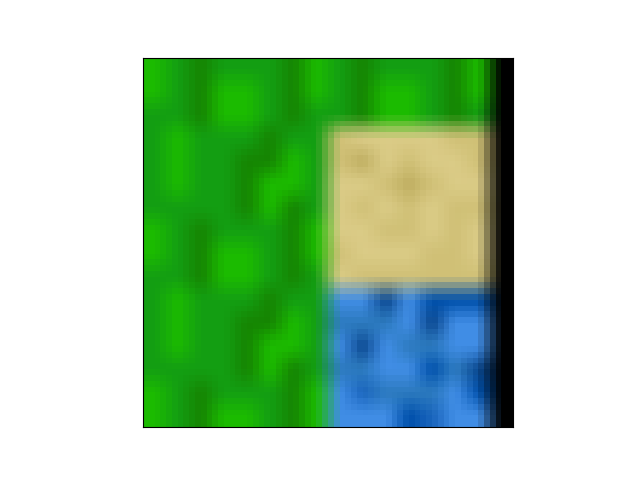} & \includegraphics[align=c, width=0.08\textwidth]{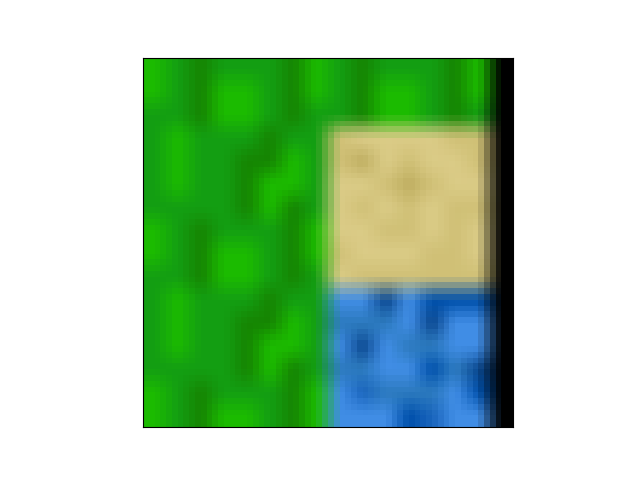} & \includegraphics[align=c, width=0.08\textwidth]{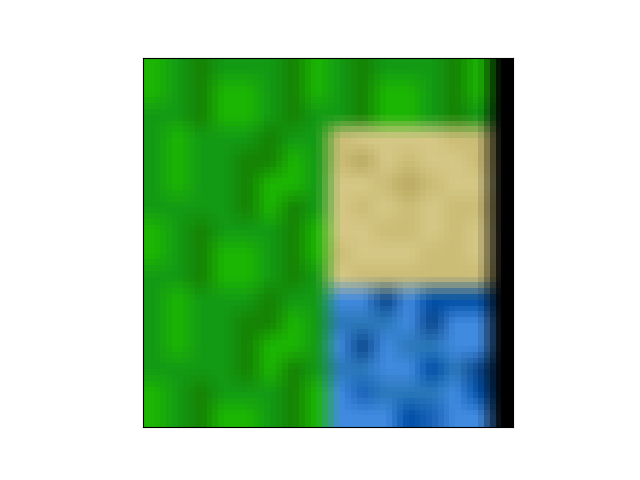} & \includegraphics[align=c, width=0.08\textwidth]{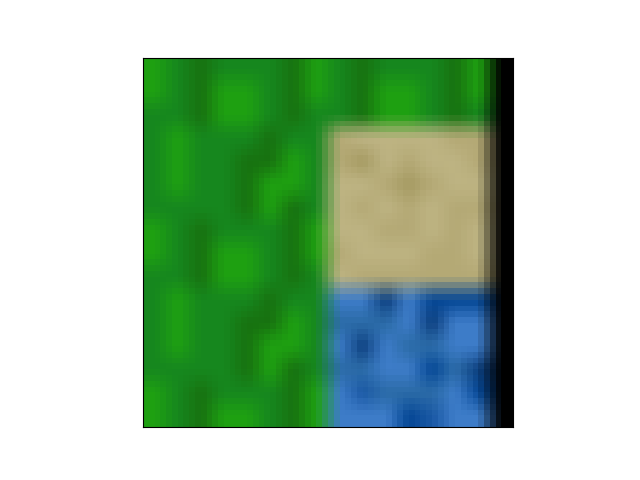} \\ 
Code 344 & 
\includegraphics[align=c, width=0.08\textwidth]{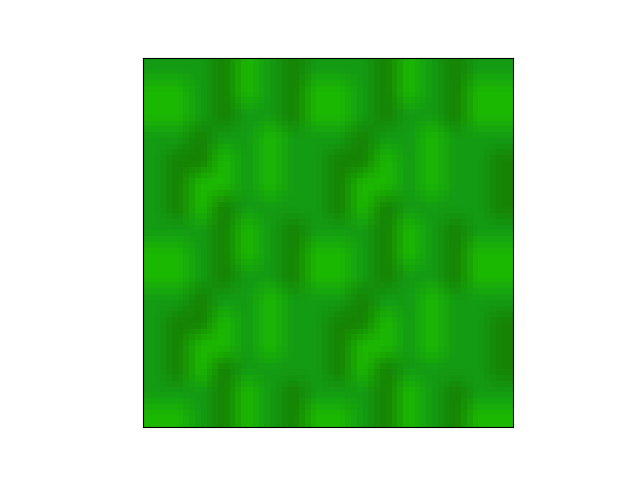} & \includegraphics[align=c, width=0.08\textwidth]{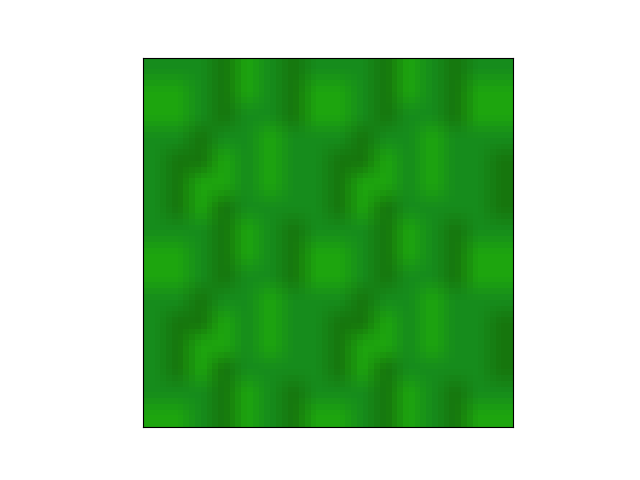} & \includegraphics[align=c, width=0.08\textwidth]{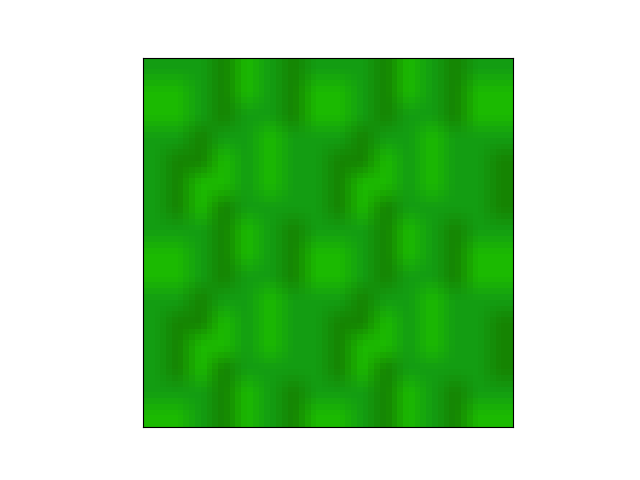} & \includegraphics[align=c, width=0.08\textwidth]{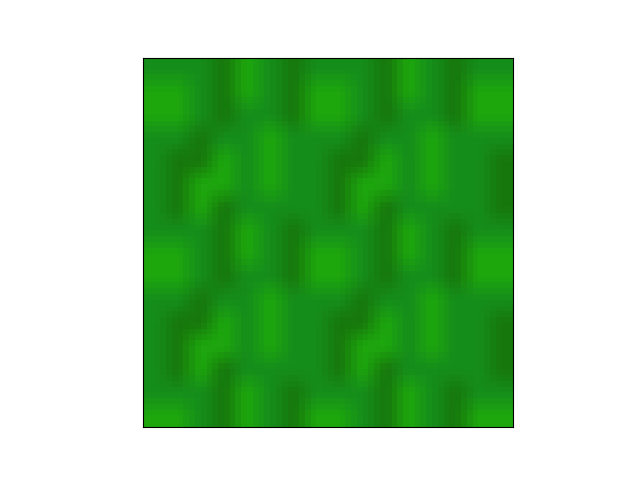} & \includegraphics[align=c, width=0.08\textwidth]{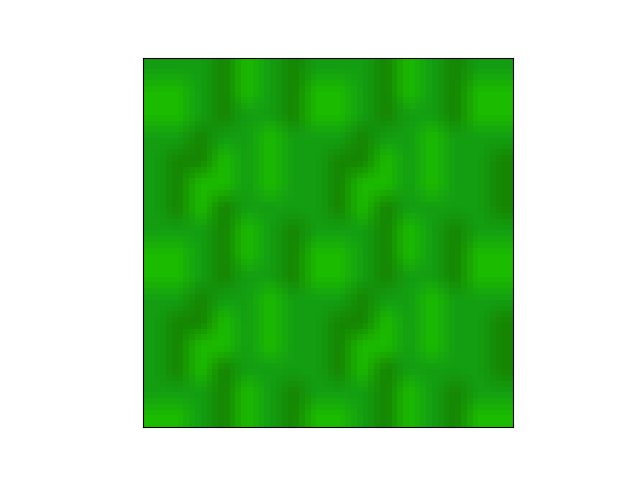} & \includegraphics[align=c, width=0.08\textwidth]{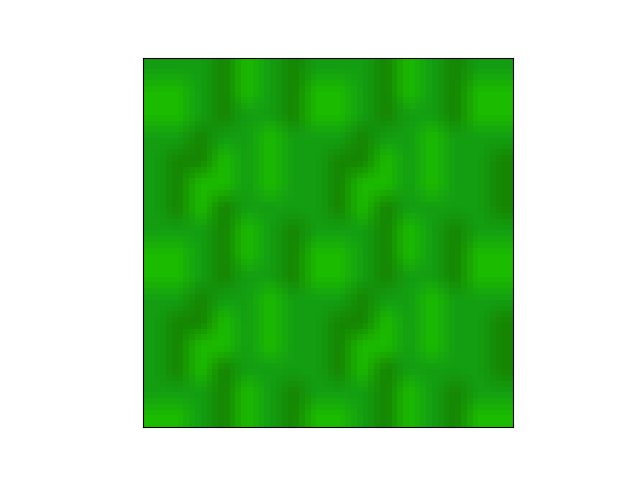} & \includegraphics[align=c, width=0.08\textwidth]{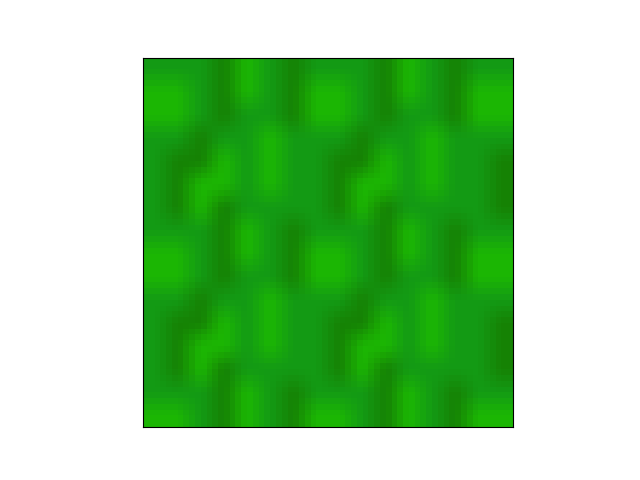} & \includegraphics[align=c, width=0.08\textwidth]{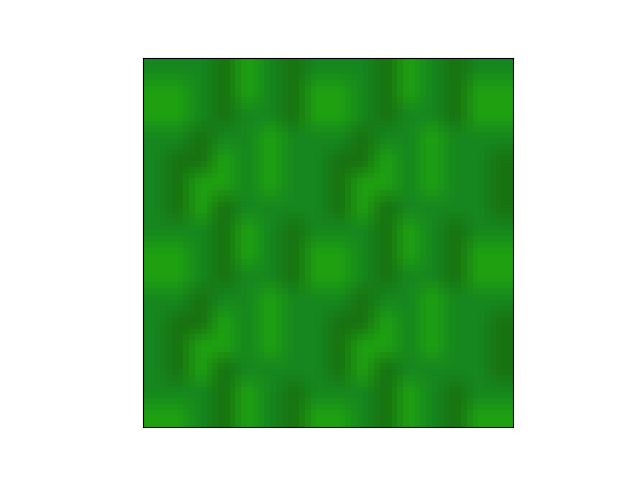} \\ 
\midrule
Code 273 & 
\includegraphics[align=c, width=0.08\textwidth]{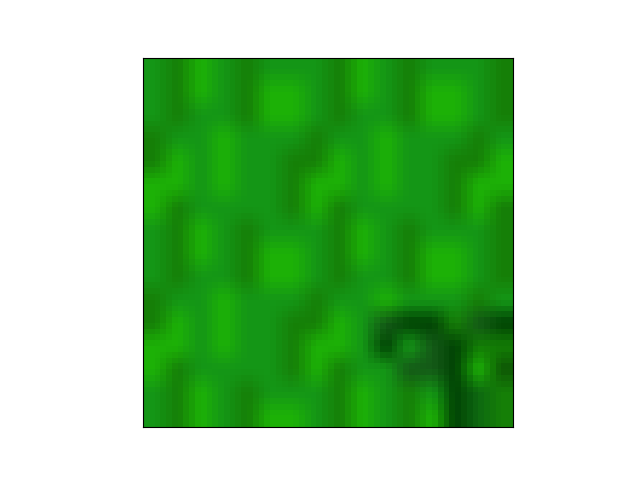} & \includegraphics[align=c, width=0.08\textwidth]{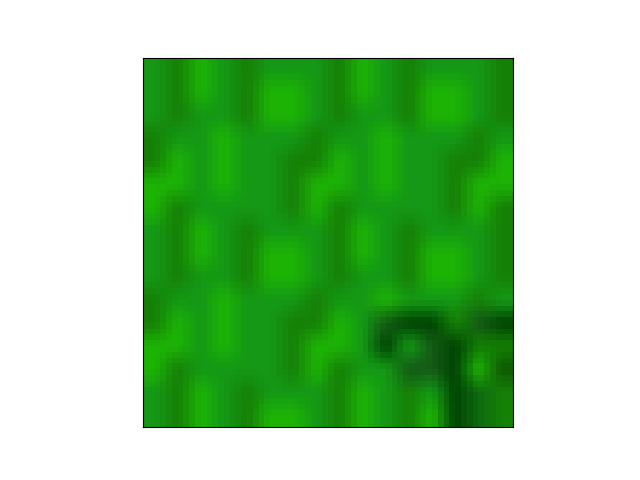} & \includegraphics[align=c, width=0.08\textwidth]{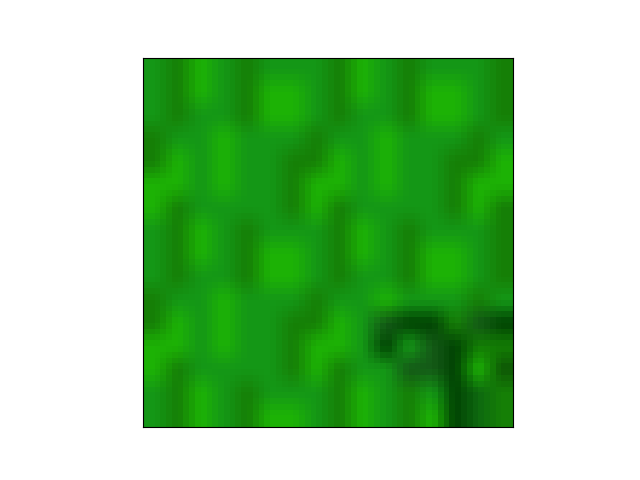} & \includegraphics[align=c, width=0.08\textwidth]{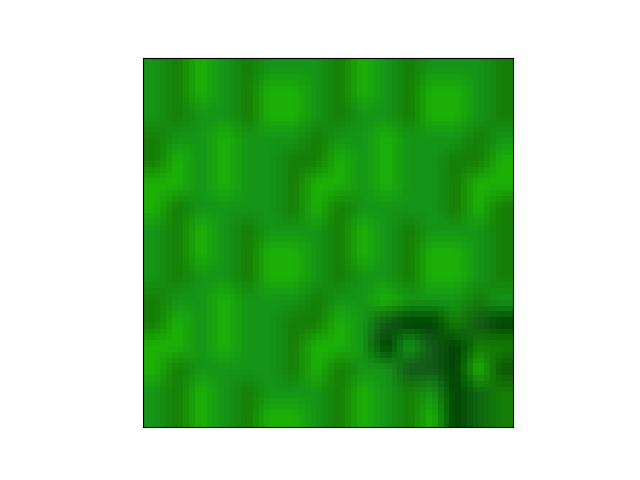} & \includegraphics[align=c, width=0.08\textwidth]{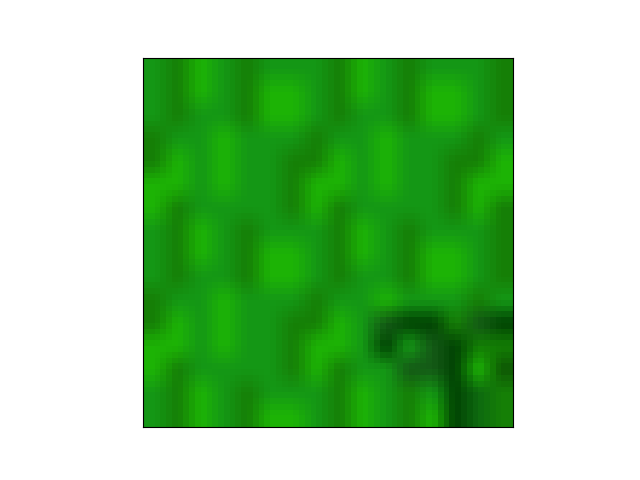} & \includegraphics[align=c, width=0.08\textwidth]{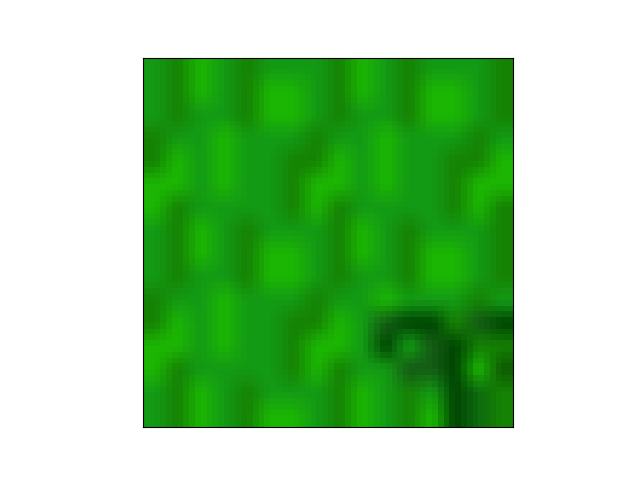} & \includegraphics[align=c, width=0.08\textwidth]{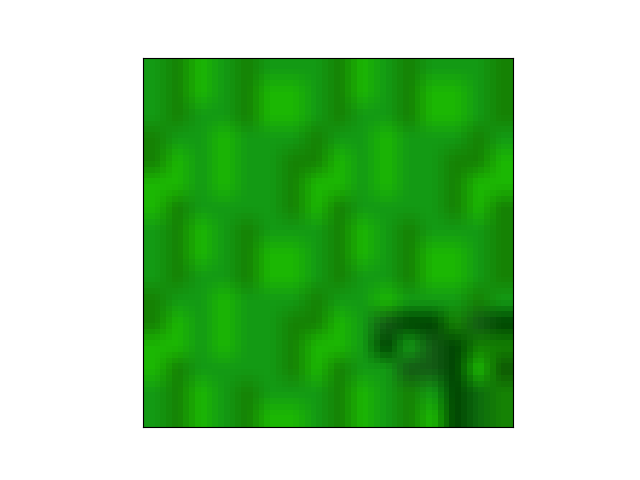} & \includegraphics[align=c, width=0.08\textwidth]{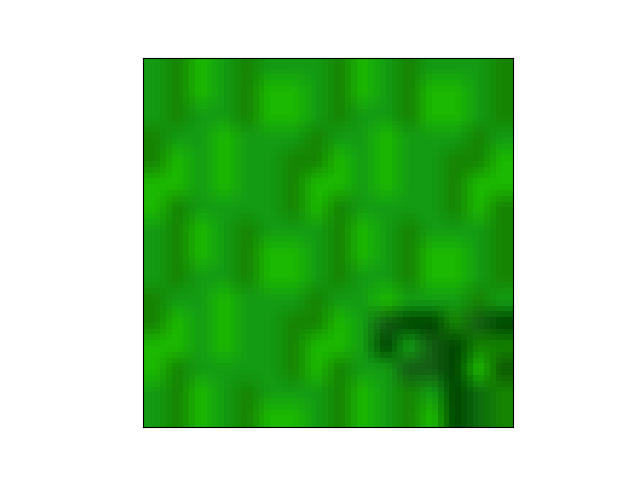} \\ 
Code 507 & 
\includegraphics[align=c, width=0.08\textwidth]{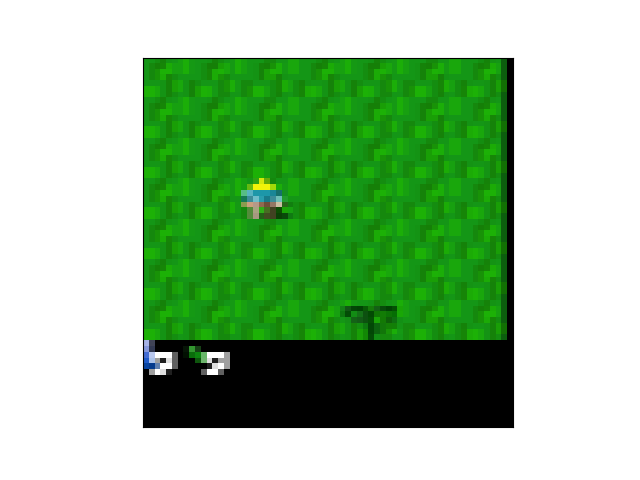} & \includegraphics[align=c, width=0.08\textwidth]{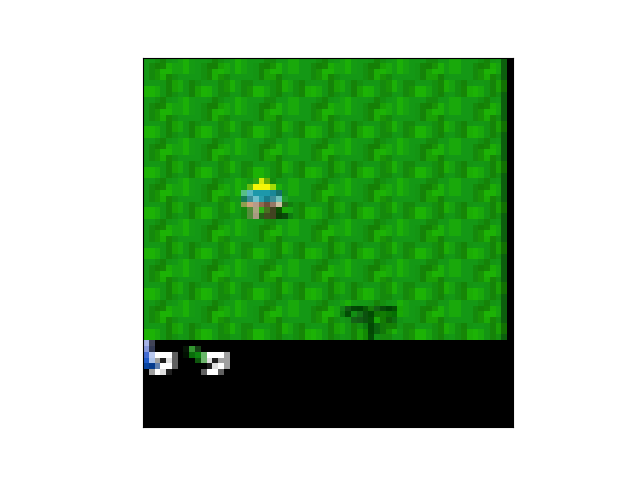} & \includegraphics[align=c, width=0.08\textwidth]{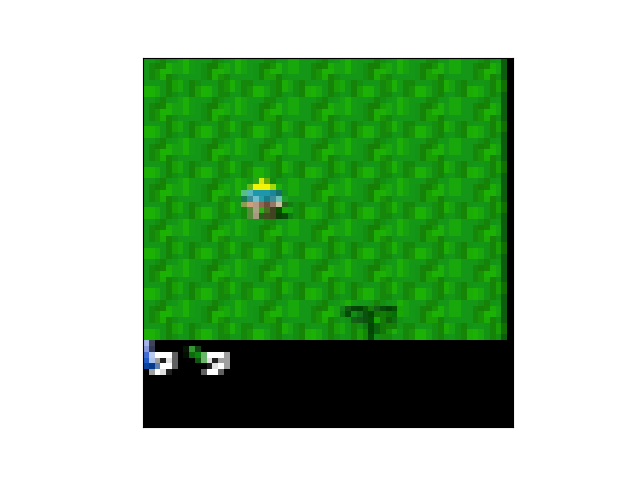} & \includegraphics[align=c, width=0.08\textwidth]{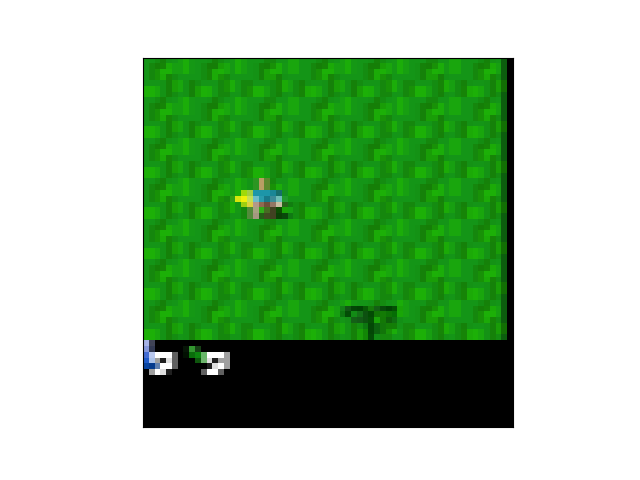} & \includegraphics[align=c, width=0.08\textwidth]{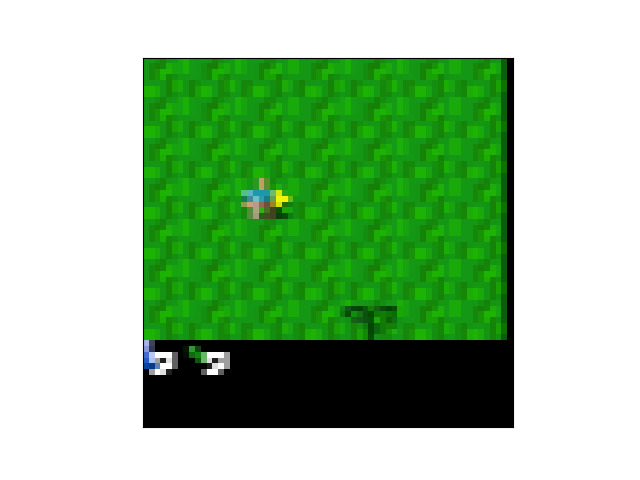} & \includegraphics[align=c, width=0.08\textwidth]{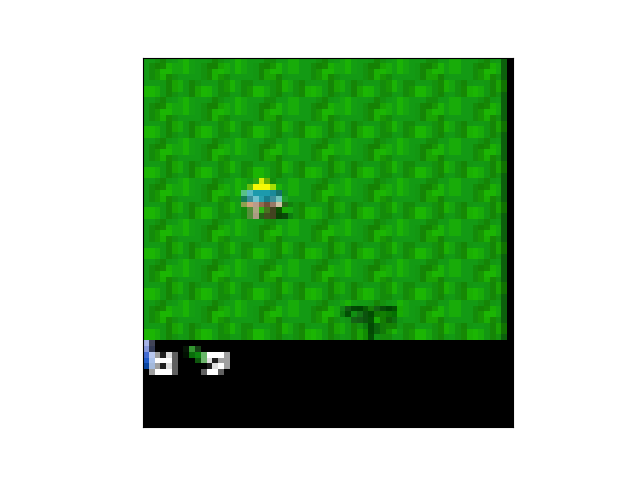} & \includegraphics[align=c, width=0.08\textwidth]{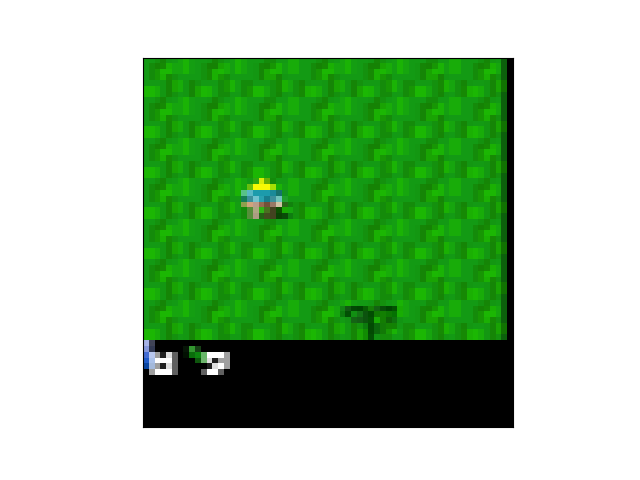} & \includegraphics[align=c, width=0.08\textwidth]{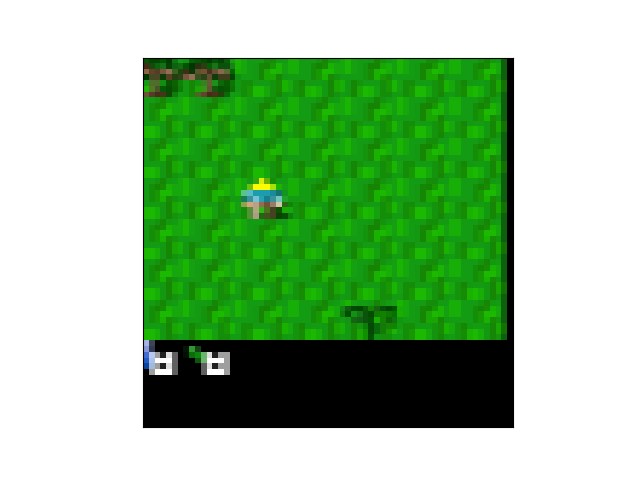} \\ 
\bottomrule 
\end{tabular}
\label{Tab:co-occur-crops}
\end{table} 

In Table \ref{Tab:co-occur-crops}, we visualize the crops from the observations where codes co-occur for the two highest co-occurring pairs, codes 344 and 424 and codes 273 and 507.
The crops reveal that each code is extremely consistent in what it captures in these instances.
Although this is a promising indication of interpretability, we believe it is still greatly limited by two factors.
The first applies only to the pair of codes 344 and 424, which is that all their instances of co-occurrence came from a single episode.
The codes selected at a given time step exhibit significant co-occurance with the codes from time steps shortly before and after them, since the observations are very similar. 
Therefore, the high consistency of the codes is less trustworthy because it is an expected behavior within an episode and not demonstrated across episodes.
The second factor limiting the impact of co-occuring codes is that, while these instances can indeed provide interpretable observations, they account for a very small portion of the entire observations collected.
Specifically, the two pairs, codes 344 and 424 and codes 273 and 507, each co-occur in roughly 0.1\% of the observations and the co-occurence rates for all codes suggest very few other pairs will provide percentages near that high.
Based on the target application of the RL system, the percent of interpretable samples needed will vary.
Given an example target of 10\%, our results suggest this may not be possible to reach by combining all the interpretable codes and co-occurring pairs without extensive analysis of all the codes and code combinations.
We argue if arduous analysis is required to reach a sufficient amount of interpretability, vector quantization by itself is not truly providing interpretability to the meaningful extent suggested by prior literature.

\end{document}